\documentclass[sigconf, nonacm]{acmart}

\newcommand\vldbpagestyle{empty} 

\usepackage[utf8]{inputenc}
\usepackage[T1]{fontenc}
\usepackage{hyperref}   
\usepackage{url}        
\usepackage{booktabs}   
\usepackage{amsfonts}   
\usepackage{nicefrac}   
\usepackage{microtype}  
\usepackage{booktabs}
\usepackage{soul} 

\usepackage{graphicx}
\usepackage[TABBOTCAP]{subfigure}
\usepackage{amsmath,amsfonts,bbm}
\usepackage{xcolor,colortbl}
\usepackage{comment}
\usepackage{multirow}
\usepackage{enumitem}

\usepackage{diagbox}
\usepackage[export]{adjustbox}
\usepackage{balance}
\usepackage{dblfloatfix}

\usepackage{array}
\newcolumntype{"}{@{\hskip\tabcolsep\vrule width 1pt\hskip\tabcolsep}}

\definecolor{LightCyan}{rgb}{0.93,1,1}
\definecolor{MedCyan}{rgb}{0.5,1,1}
\definecolor{DarkCyan}{rgb}{0.0, 0.55, 0.55}

\newcommand{\minipa}[1]{\vspace{0.02in} \noindent \textbf{#1}}
\newcommand{\minipb}[1]{\vspace{0.02in} \noindent \underline{#1}}
\newcommand{\ssection}[1]{\vspace{-3mm}\section{#1}\vspace{-0.5mm}}
\newcommand{\ssubsection}[1]{\vspace{-2mm}\subsection{#1}\vspace{-0.5mm}}
\newcommand{\system}{Exathlon}
\newcommand{\msbold}[1]{\boldsymbol{~#1}}

\begin{document}
\title{Exathlon: A Benchmark for Explainable Anomaly Detection \\ over Time Series}
\author{
Vincent Jacob$^\dagger$
\hbox{~~~~}Fei Song$^\dagger$ 
\hbox{~~~~}Arnaud Stiegler$^\dagger$ 
\hbox{~~~~}Bijan Rad$^\dagger$ 
\hbox{~~~~}Yanlei Diao$^\dagger$ 
\hbox{~~~~}Nesime Tatbul$^\ast$
}
\affiliation{\institution{
$^\dagger$Ecole Polytechnique, France
\hbox{~~~~}$^\ast$Intel Labs and MIT, USA}
}
\email{
{vincent.jacob, fei.song, arnaud.stiegler, bijan.rad, yanlei.diao}@polytechnique.edu, tatbul@csail.mit.edu
}

\begin{abstract}
Access to high-quality data repositories and benchmarks have been instrumental in advancing the state of the art in many experimental research domains.
While advanced analytics tasks over time series data have been gaining lots of attention, lack of such community resources severely limits scientific progress.
In this paper, we present \system, the first comprehensive public benchmark for explainable anomaly detection over high-dimensional time series data. \system\ has been systematically constructed based on real data traces from repeated executions of large-scale stream processing jobs on an Apache Spark cluster. Some of these executions were intentionally disturbed by introducing instances of six different types of anomalous events (e.g., misbehaving inputs, resource contention, process failures). For each of the anomaly instances, ground truth labels for the root cause interval as well as those for the extended effect interval are provided, supporting the development and evaluation of a wide range of anomaly detection (AD) and explanation discovery (ED) tasks.
We demonstrate the practical utility of \system's dataset, evaluation methodology, and end-to-end data science pipeline design through an experimental study with three state-of-the-art AD and ED techniques.
\end{abstract}

\maketitle

\pagestyle{\vldbpagestyle}

\ssection{Introduction} 
\label{sec:intro}

Time series is one of the most ubiquitous types of data in our increasingly digital and connected society. Advanced analytics capabilities such as detecting anomalies and explaining them are crucial in understanding and reacting to temporal phenomena captured by this rich data type. {\em Anomaly detection} (AD) refers to the task of identifying patterns in data that deviate from a given notion of normal behavior \cite{ChandolaBK2009}. It finds use in almost every domain where data is plenty, but unusual patterns are the most critical to respond (e.g., cloud telemetry, autonomous driving, financial fraud management). AD over time series data has been of particular interest, not only because time-oriented data is highly prevalent and voluminous, but also more challenging to analyze due to its complex and diverse nature: multivariate time series can consist of 1000s of dimensions; anomalous patterns may be of arbitrary length and shape; there may be intricate cause and effect relationships among these patterns; data is rarely clean. Furthermore, by helping uncover how or why a detected anomaly may have happened, {\em explanation discovery} (ED) forms a crucial capability for any time series AD system.

Recent advances in data science and machine learning (ML) significantly reinforced the need for developing robust anomaly detection and explanation solutions that can be reliably deployed in production environments \cite{explainit:sigmod19, ma:pvldb20}. However, progress has been rather slow and limited. While there is extensive research activity going on \cite{DLAD-survey}, proposed solutions have been mostly adhoc and far from being generalizable to realistic settings. We believe that one of the critical roadblocks to progress has been the lack of open data repositories and benchmarks to serve as a common ground for reproducible research and experimentation. Indeed, access to such community resources has been instrumental in advancing the state of the art in many other domains (e.g., \cite{imagenet:cvpr09, objectnet:neurips19}). Inspired by those efforts, in this paper, we propose {\em \system, the first comprehensive public benchmark for explainable anomaly detection over high-dimensional time series data}.

\system\ focuses on the familiar domain of metric monitoring in large-scale computing systems, and provides a benchmarking platform that consists of:
(i) a curated anomaly dataset,
(ii) a novel benchmarking methodology for AD and ED, and
(iii) an end-to-end data science pipeline for implementing and evaluating AD and ED algorithms based on the provided dataset and methodology. More specifically, we make the following contributions in this work:

\noindent
{\bf Dataset.}
We constructed \system\ systematically based on real data traces collected from around 100 repeated executions of 10 distributed streaming jobs on a Spark cluster over 2.5 months. Inspired by  chaos engineering in industry~\cite{Chaos2016}, our traces were obtained by disturbing more than 30  job executions with nearly 100 instances of 6 different classes of anomalous events (e.g., misbehaving inputs, resource contention, process failures). For each of these anomalies, we provide ground truth labels for both the root cause interval and the corresponding effect interval,  enabling the use of our dataset in a wide range of AD and ED tasks. Overall, both the normal ({\em undisturbed}) and anomalous ({\em disturbed}) traces contain enough variety (including some noise due to Spark's inherent behavior) to capture real-world data characteristics in this domain (Table \ref{tab:dataset}).

\noindent
{\bf Evaluation Methodology.}
\system\ evaluates AD and ED algorithms in terms of two orthogonal aspects: functionality and computational performance.
For AD,
we primarily target {\em semi-supervised techniques} (i.e., with a model developed/trained using only normal data, possibly with occasional noise, and then tested against anomalous data) for {\em range-based anomalies} (i.e., contextual and collective anomalies occurring over a time interval instead of only at a single time point \cite{ChandolaBK2009}) over {\em high-dimensional time series} (i.e., multivariate with 1000s of dimensions). This decision is informed by our observation of this being the most common and inclusive usage scenario in practice.
For ED,
we broadly consider both {\em model-free} (e.g., \cite{macrobase:sigmod17, ZDM2017}) and {\em model-dependent} (e.g., \cite{Ribeiro0G16}) techniques.
AD functionality is evaluated under four well-defined model learning settings, based on four key evaluation criteria -- anomaly existence, range detection, early detection, and exactly-once detection -- using a novel range-based accuracy framework \cite{TatbulLZAG18}. Similarly, ED functionality is tested for two capabilities -- local explanation and global explanation -- each measured in terms of conciseness, consistency, and accuracy. Computational efficiency and scalability for both AD model training/inference as well as for ED execution can also be evaluated at varying data dimensions and sampling rates.
Overall, \system\ provides a rich and challenging testbed with a well-organized evaluation methodology (Table \ref{tab:benchmark-design}).

\begin{figure}[t]
	\begin{center}
		\leavevmode
		\includegraphics[width=1\columnwidth,height=3.7cm]{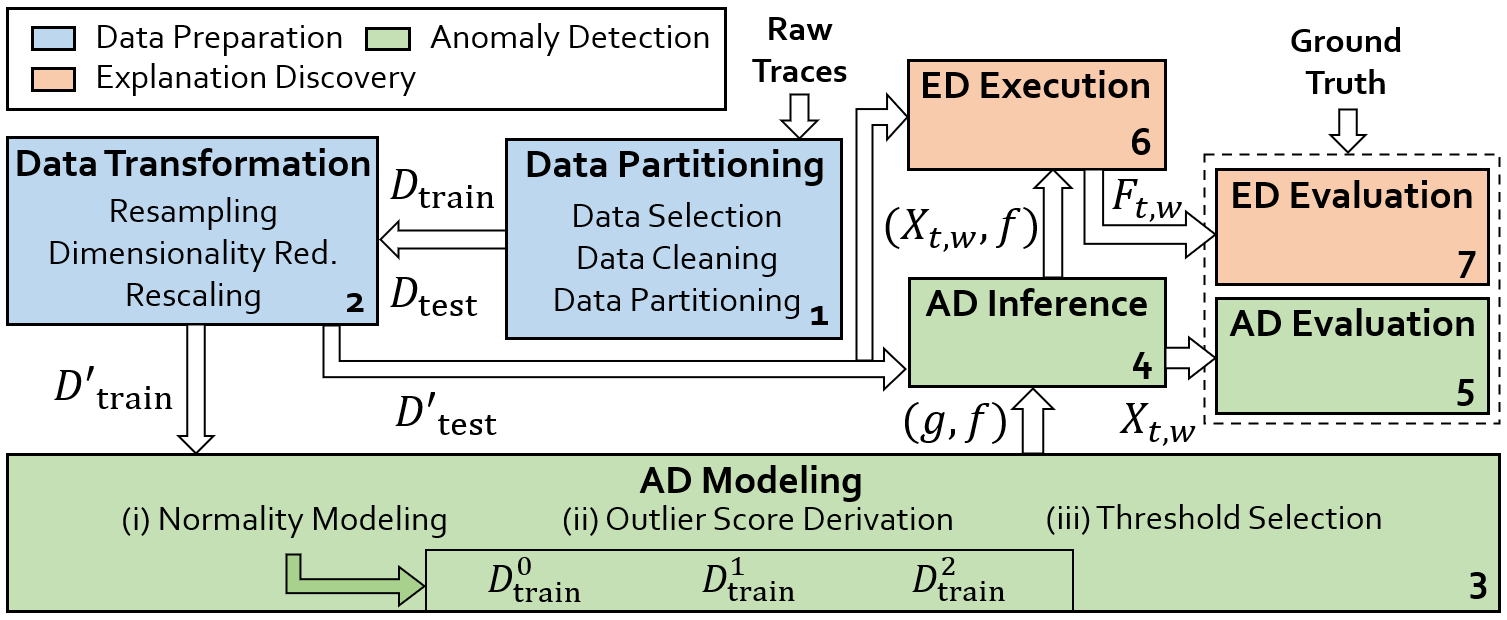}
	\end{center}
	\caption{\small A pipeline for explainable AD on multivariate time series}
	\vspace{-0.1in}
    	\label{fig:pipeline}
\end{figure}

\noindent
{\bf Data Science Pipeline.}
We designed an end-to-end pipeline for explainable time series anomaly detection. This pipeline includes all the data processing steps necessary to turn our raw datasets into AD and ED results together with their benchmark scores. Our design is modular and extensible. This not only makes it easy to implement new AD and ED techniques to benchmark, but also allows creating multiple variants of pipeline steps to experiment with and compare. For example, training data preparation for different AD learning settings or scoring AD results for different criteria levels can be easily configured, run, and compared in our pipeline (Figure \ref{fig:pipeline}).

\noindent
{\bf Experimental Study.}
We provide the first experimental study evaluating a representative set of state-of-the-art AD and ED techniques to illustrate the usage and benefits of \system. Results suggest that our dataset carries useful signals that can be picked up by the tested AD and ED algorithms in a way that can be effectively quantified by our evaluation criteria and metrics. Furthermore, we observe that our benchmarking framework exposes increasing levels of challenges to stress-test these algorithms in a systematic way (\S \ref{sec:experiments}).

Compared to current public resources for time series AD research~\cite{uea-dataset, ucr-dataset, uci-dataset, numenta:icmla15, odds-library}, a key contribution of \system\ is that it comprehensively covers one challenging application domain end to end, as opposed to providing multiple smaller and simpler datasets from several independent domains. Furthermore, a public benchmark for time series ED research with ground truth labels is largely lacking today, making it hard to evaluate and compare an increasing number of published papers on this important topic. Thus, we believe \system\ provides an opportunity for a more in-depth investigation and evaluation of models and algorithms in both time series AD and ED, potentially revealing new insights for accelerating research progress in explainable anomaly detection. 

In the rest of the paper, we first briefly summarize related work. After presenting our dataset, evaluation methodology, and 
pipeline design in more detail, we show the practical utility of \system\ through an experimental analysis of three state-of-the-art AD and ED algorithms using a selected set of evaluation criteria and settings from our benchmark.
Finally, we conclude with an outline of future directions.
The dataset, code, and documentation for \system\ are publicly available at \url{https://github.com/exathlonbenchmark/exathlon}.

\ssection{Related Work} 
\label{sec:related}

\noindent
{\bf Datasets and Benchmarks.}
Benchmarks to evaluate database (DB) system performance have been around for more than 30 years \cite{bench-handbook, tpc-bench, spec-bench}.
In addition to industry-standard benchmarks for relational DB workloads such as TPC-C and TPC-H, new domain-specific benchmarks for emerging workloads have been proposed (e.g., Linear Road Benchmark for stream processing \cite{linear-road}, YCSB for scalable key-value stores \cite{ycsb}, BigBench for big data analytics \cite{bigbench}). The main focus of these benchmarks has been on computational performance. With recent benchmarks for ML/DL-based advanced data analytics such as ADABench and DAWNBench, there has been a focus shift toward end-to-end ML pipelines and new evaluation metrics such as time to accuracy \cite{adabench, dawnbench}. Like in DB and systems communities, the ML community has also been publishing datasets and benchmarks to support research in many problem domains from object recognition to natural language processing \cite{imagenet:cvpr09, objectnet:neurips19, wordnet:cacm95, glue:iclr19}. Well-known data archives for time series research include: UCI \cite{uci-dataset}, UCR \cite{ucr-dataset}, and UEA \cite{uea-dataset}. These archives provide real-world data collections created for general ML tasks, such as classification and clustering. While the need for systematically constructing AD benchmarks from real data has also been recognized by others \cite{emmott:odd13}, public availability of anomaly datasets is still limited \cite{odds-library}. To our knowledge, Numenta Anomaly Benchmark (NAB) is the only public benchmark designed for time series AD \cite{numenta:icmla15}. NAB provides 50+ real and artificial datasets, primarily focusing on real-time AD for streaming data. Compared to ours, each of these datasets is much smaller in scale and dimensionality, and does not capture any information to enable ED. NAB also has several technical weaknesses that hinder its use in practice (e.g., ambiguities in its scoring function, missing values in its datasets) \cite{singh:ijcnn17}.

\noindent
{\bf Anomaly Detection (AD).} There is a long history of research in AD \cite{ChandolaBK2009, GuptaGA2014}. The high degree of diversity in data characteristics, anomaly types, and application domains  has led to a plethora of AD approaches from simple statistical methods~\cite{bianco01} 
to distance-based~\cite{Tran2015}, density-based~\cite{ChandolaBK2009,LOF}, and isolation forests~\cite{IsolationForest} to deep learning (DL) methods~\cite{DLAD-survey}. 
It is beyond the scope of this paper to provide a complete survey; we refer the reader to recent survey papers for a full discussion of such methods ~\cite{ChandolaBK2009,GuptaGA2014,DLAD-survey}. 
In our experimental study, we particularly focus on three DL methods that represent the recent state of the art
~\cite{MVSA, XuCZLBLLZPFCWQ18, SchleglSWSL17} (detailed in \S \ref{sec:experiments}). Such DL methods have the potential 
to handle a variety of anomaly patterns, such as contextual and collective patterns \cite{ChandolaBK2009}, and overcome known limitations of previous 
density- and distance-based methods that are very sensitive to data dimensions.

\noindent
{\bf Interpretable Machine Learning.}
Interpretable ML has recently attracted a lot of attention
~\cite{Molnar-book}. 
Relevant techniques generally belong to two broad families: {\em interpretable models} and {\em model-agnostic methods}.
Interpretable models 
directly build a human-readable model from the data (e.g., linear or logistic regression, decision trees or rules) \cite{Molnar-book}.
In contrast, model-agnostic methods separate explanations from the ML model, hence offering the flexibility to mix and match ML models with interpretation methods.
In the model-agnostic family, several methods obtain interpretable classifiers by perturbing the inputs and observing the response~\cite{LakkarajuBL16,Ribeiro0G16,Ribeiro0G18,SundararajanTY17}. 
LIME~\cite{Ribeiro0G16} explains a  prediction of any classifier by approximating it locally with an interpretable sparse linear model, and explains the overall model by selecting a set of representative instances with explanations. As such, it  is generally considered a method for {\em local explanations}. We evaluate LIME in our experimental study.
Anchors~\cite{Ribeiro0G18} improved upon LIME by replacing its linear model with a logical rule for explaining a data instance. It offers better coverage of data points in a local neighborhood, but does not support time series data.
SHAP scores~\cite{LundbergL17}, RESP scores~\cite{BertossiLSSV20}, and axiomatic attribution~\cite{SundararajanTY17} are also instance-level explanations that assign a numerical score to each feature, representing their importance in the outcome.
In contrast to local explanations,  other  work aims to explain a model via {\em global explanations}. 
Some of them approximate a DL model using a decision tree~\cite{FrosstH17,WuHPZ2018}, or by learning a decision set~\cite{LakkarajuBL16,kopp:eswa20} directly as explainable models.
All of these methods suffer from lacking a benchmark dataset and evaluation methodology. The FICO challenge was designed to evaluate such methods using a home loan application dataset, with a known label (high or low risk) for each application, but it relies on manual evaluation of returned explanations by real-world data scientists ~\cite{FICOdata}. As a result, it remains hard to compare different ED methods due to the lack of ground truth explanations and automated evaluation procedures. 

\noindent
{\bf Explaining Outliers in Data Streams.}
There is a handful of work in explaining outliers in data streams.
Given normal and abnormal time periods by the user,
EXstream finds explanations to best distinguish the abnormal periods from the normal ones ~\cite{ZDM2017}. 
MacroBase  helps the user prioritize attention over data streams, with modules for both AD and  ED tasks~\cite{macrobase:sigmod17}. Its AD module uses simple statistical methods like MAD, which is known to be suitable only for detecting simple point outliers~\cite{ChandolaBK2009}.
For a detected anomaly, MacroBase's ED module discovers an explanation in the form of conjunctive predicates, by using a frequent itemset mining framework that takes minimum support and risk ratio as input parameters.
We evaluate both of these techniques in our experimental study.

\noindent
\textbf{Explaining Outliers in SQL Query Results.}
Scorpion explains outliers in group-by aggregate queries by searching through various subsets of the tuples that were used to compute the query answers~\cite{Wu2013scorpion}. 
Given a set of explanation templates by the user, Roy et al.'s approach performs precomputation in a given DB to enable interactive ED \cite{Roy2015explaining}.
Similarly, given a table, El Gebaly et al.'s work constructs an explanation table and finds patterns affecting a binary value of each tuple ~\cite{el2014interpretable}.
These approaches target traditional DB workloads and are not applicable to our problem.
There have been recent industrial efforts on time series anomaly explanation and root cause analysis in DB systems \cite{explainit:sigmod19, ma:pvldb20}. These approaches require a variety of inputs from the user, e.g., causal hypotheses \cite{explainit:sigmod19} or labels of root causes~\cite{ma:pvldb20}, whereas our work focuses on semi-supervised learning for explainable AD. Moreover, these systems are largely based on proprietary code and datasets that are not accessible to the research community.

\begin{figure*}[t]
	\centering
	\begin{tabular}{lcc}
		\hspace{-0.1in}
		\subfigure[\small{Spark application monitoring}]
		{\label{fig:app}\includegraphics[width=0.33\textwidth,height=3.3cm]{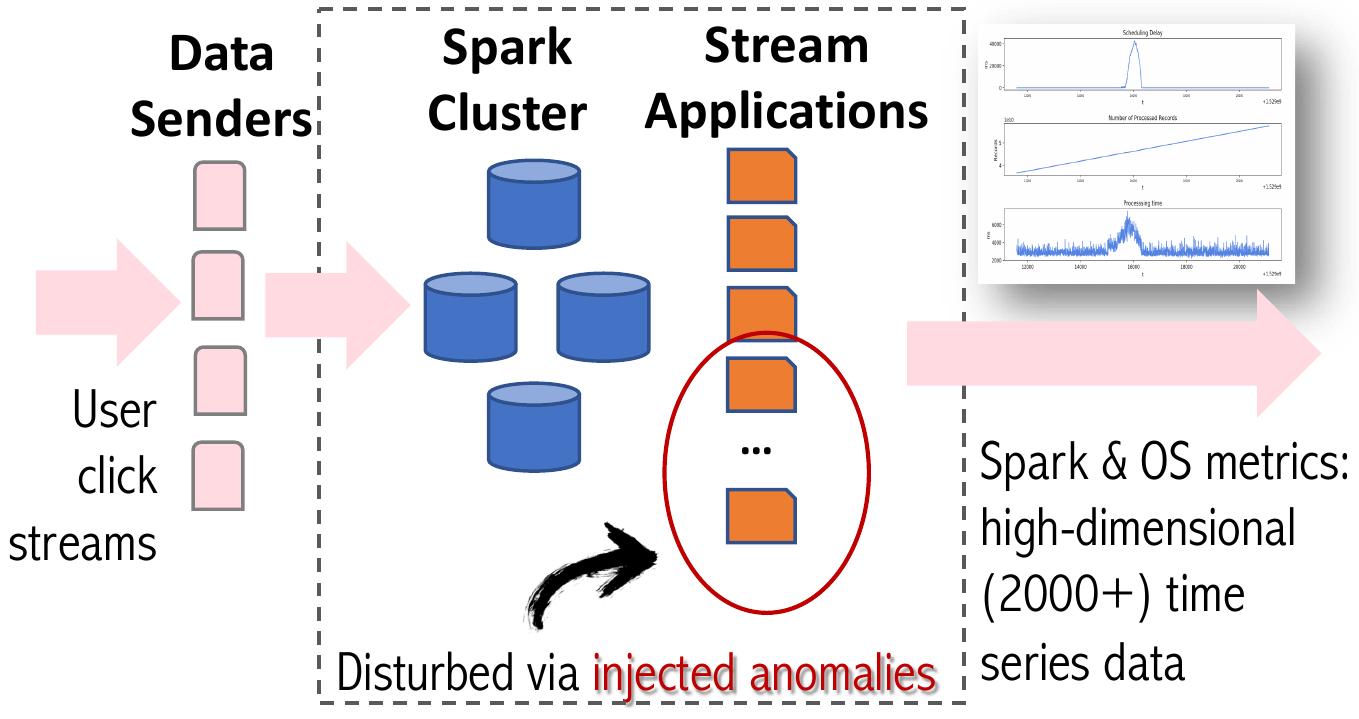}}
		& 
		\subfigure[\small{Spark execution environment}]
		{\label{fig:spark}\includegraphics[width=0.29\textwidth,height=3.3cm]{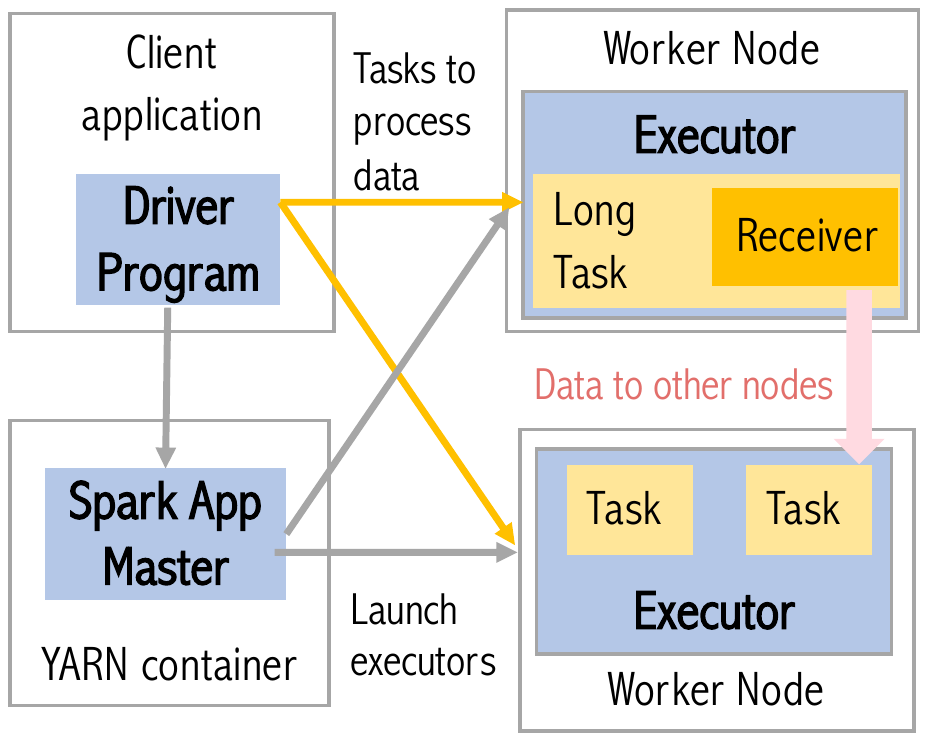}}
		&
		\hspace{-0.1in}
		\subfigure[\small{Trace with bursty input anomalies}]
		{\label{fig:burstyinput}\includegraphics[width=0.33\textwidth,height=3.3cm]{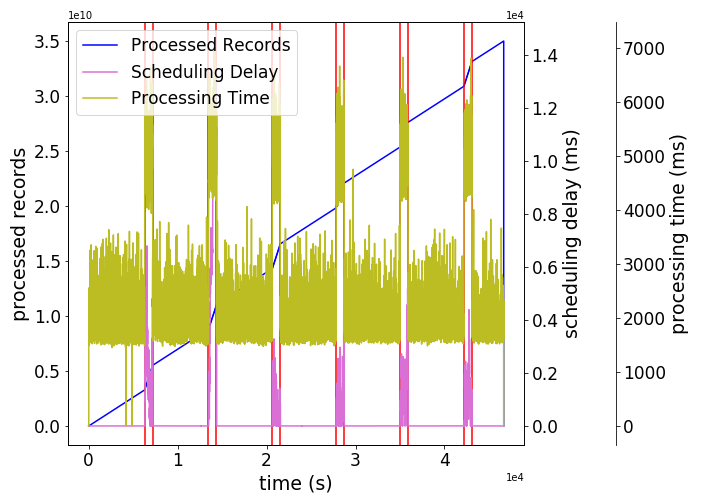}}
	\end{tabular}
\vspace{-0.2in}
\caption{\small Spark application monitoring, and metrics observed in anomaly instances (a pair of red vertical bars marks a root cause event) }
\label{fig:all-normal}
\vspace{-0.1in}
\end{figure*}

\begin{table*}[t]
\centering
{\footnotesize
\subfigure[Metrics and data size]
{
\adjustbox{valign=b}{
\setlength\tabcolsep{2pt}
\begin{tabular}{|r"c|c|c|}
\hline
{\bf Metric}    & Spark UI & Spark UI & OS     \\
{\bf Type}      & Driver   & Executor & (Nmon) \\
\hline
{\bf \# of}     &         & 5 x 140  & 4 x 335 \\
{\bf Metrics}   & 243     & = 700    & = 1340  \\
\hline
{\bf Total}     & \multicolumn{3}{c|}{2,283} \\
\hline \hline
{\bf Frequency} & \multicolumn{3}{c|}{1 data item per second} \\
\hline
{\bf Data Items}  & \multicolumn{3}{c|}{2,335,781} \\
\hline
{\bf Duration}      & \multicolumn{3}{c|}{649 hours} \\
\hline
{\bf Total Size}       & \multicolumn{3}{c|}{24.6 GB} \\
\hline
\end{tabular}
\label{tab:metrics}
}}
\subfigure[Undisturbed traces, disturbed traces,  and ground truth labels of 97 anomalies ]
{
\adjustbox{valign=b}{
\setlength\tabcolsep{2pt}
\begin{tabular}{|c|l|c|c|c|c|}
\hline
{\bf Trace} & \multicolumn{1}{c|}{{\bf Anomaly}} & {\bf \# of}     & {\bf Anomaly} & {\bf Anomaly Length (RCI + EEI)}  & {\bf Data} \\
{\bf Type}  & \multicolumn{1}{c|}{{\bf Type}}    & {\bf Traces} & {\bf Instances}  & {\bf min, avg, max} & {\bf Items}  \\
\hline
Undisturbed    & \multicolumn{1}{c|}{N/A}    & 59 & N/A & N/A  & 1.4M  \\ %
\hline
Disturbed & T1: Bursty input & 6  & 29  & 15m, 22m, 33m & 360K  \\
\hline
Disturbed & T2: Bursty input until crash & 7  & 7   & 8m, 35m, 1.5h  & 31K  \\
\hline
Disturbed & T3: Stalled input & 4  & 16  & 14m, 16m, 16m    & 187K  \\
\hline
Disturbed & T4: CPU contention & 6  & 26  & 8m, 15m, 27m      & 181K \\
\hline
Disturbed & T5: Driver failure & \multirow{2}{*}{11}    & 9   &  1m, 1m, 1m & \multirow{2}{*}{128K}  \\
\cline{1-2}\cline{4-5}
Disturbed & T6: Executor failure &    & 10   &  2m, 23m, 2.8h &  \\
\hline \hline
Ground & \multicolumn{5}{l|}{(\texttt{app\_id}, \texttt{trace\_id}, \texttt{anomaly\_type}, \texttt{root\_cause\_start}, \texttt{root\_cause\_end},} \\
 Truth  & \multicolumn{5}{l|}{\, \texttt{extended\_effect\_start}, \texttt{extended\_effect\_end})}
\\ \hline
\end{tabular}
\label{tab:traces}
}
}
}
\vspace{-0.1in}
\caption{\small The \system\ dataset} %
\label{tab:dataset}
\vspace{-0.3in}
\end{table*}

\ssection{Dataset} \label{sec:anomalies}

The \system\ dataset has been systematically constructed based on real data traces collected from a use case scenario that we implemented on Apache Spark. In this section, we first describe this scenario, followed by the details of how we created the normal and anomalous data traces themselves.

\ssubsection{Use Case: Spark Application Monitoring}

Large-scale data analytics applications are deployed on Apache Spark clusters everyday. Monitoring the execution of these jobs to ensure their correct and timely completion via AD can be business-critical. For example, some of the largest e-commerce platforms run Spark jobs on petabytes of data each day to analyze purchase patterns, target offers, and enhance customer experiences~\cite{Spark-uses}. Since results of these jobs affect immediate business decisions such as inventory management and sales strategies, they are often specified with deadlines. Anomalies that occur in job execution prevent analytical jobs from meeting their deadlines and hence cause disruption to critical operations on those e-commerce platforms. 
We model this widespread and challenging AD use case in our benchmark.

\noindent
{\bf System Setup.} 
Our Spark workload consists of 10 stream processing applications (see Appendix~\ref{appendix:spark-applications}), analyzing user click streams from the WorldCup 1998 website~\cite{LiMD+12}.
As in Figure~\ref{fig:app}, {\em Data Sender} servers send 
streams at a controlled {\em input rate} to a Spark cluster of 4 nodes, each with 2 Intel{\scriptsize{\textsuperscript{\textregistered}}} Xeon{\scriptsize{\textsuperscript{\textregistered}}} Gold 6130 16-core processors, 768GB of memory, and 64TB disk. Each application has certain workload characteristics (e.g., CPU or I/O intensive) and is executed by Spark in a distributed manner, as in Figure~\ref{fig:spark}.
Submitted an application, Spark launches a {\em Driver} process to coordinate the execution. The driver connects to a resource manager (Apache Hadoop YARN), which launches {\em Executor} processes on a subset of 
nodes where tasks (units of work on a data partition, e.g., \texttt{map} or \texttt{reduce}) will be executed 
in parallel. 
Given 32 cores, each node can also run tasks from multiple applications concurrently. As common real-world practice, we run 5/10 randomly selected applications at a time. 
The placement of Driver and Executor processes to cluster nodes is decided by YARN based on data locality, load on nodes, etc. Except for I/O activities, YARN offers container isolation for resource usage of all parallel processes.

\noindent
{\bf Trace Collection.} We ran the 10 Spark streaming applications in our 4-node cluster over a 2.5-month period. The data collected from each run of a Spark streaming application is called a {\em Trace}.
Some of the traces were manually pruned, because they were affected by cluster downtimes or the injected anomalies were not well reflected in the data due to failed attempts.
After this manual pruning, we kept 93 traces to constitute the \system\ dataset.

\noindent
{\bf Metrics Collected.}
During each application execution, we collected metrics from both the Spark Monitoring and Instrumentation Interface (UI) and underlying operating system (OS).
Table \ref{tab:metrics} gives a summary of the metrics collected per trace. The Driver offers 243 Spark UI metrics covering scheduling delay, statistics on the streaming data received and processed, etc. 
Each executor provides 140 metrics on various time measurements, data sizes, network traffic, as well as memory and I/O activities. As we wanted to keep the number of metrics the same for all traces, we set a fixed limit of 5 for the number of Spark executors (3 active + 2 backup). This way, even if an active executor fails during a run and a backup takes over, the number of metrics collected stays the same, $5 \times 140 = 700$, with null values set for inactive executors. 
335 OS metrics for each of the 4 cluster nodes 
were collected using the \texttt{Nmon} command, capturing CPU time, network traffic, memory usage, etc. 
All in all, each trace consists of a total of 2,283 metrics recorded each second for 7 hours on average, constituting a multi-dimensional time series.

\ssubsection{Undisturbed vs. Disturbed Traces}

In generating our traces, we followed an approach similar to chaos engineering (i.e., an approach devised by high-tech companies like Netflix for injecting failures and workload surges into a production system to verify/improve its reliability) ~\cite{Chaos2016} and general systems monitoring (e.g., Microsoft's NetMedic \cite{netmedic}). Thus, we first generated {\em undisturbed traces} to characterize the normal execution behavior of our Spark cluster; we then introduced various anomalous events to generate {\em disturbed traces}. Table \ref{tab:traces} provides an overview.

\noindent
{\bf Undisturbed Traces.}
Uninterrupted executions of 5/10 randomly selected applications at a time, at parameter settings within the capacity limits of our Spark cluster, over a period of 1 month, gave us 59 normal traces of 15.3GB in size. Any instances of occasional cluster downtime were manually removed from these traces. It is important to note that, although undisturbed, these traces still exhibit occasional variations in metrics due to Spark's inherent system mechanisms (e.g., checkpointing, CPU usage by a DataNode in the distributed file system). Since such variations do appear in almost every trace, we consider them as part of the normal system behavior. In other words, our normal data traces include some ``noise'', as most real-world datasets typically do.

\begin{table*}[t]
\setlength{\textfloatsep}{-1ex}
\setlength{\abovecaptionskip}{-1ex}
\setlength{\belowcaptionskip}{1ex}
\centering
{\small
\begin{tabular}{|r|l|l|l|}
\hline
& \multicolumn{1}{c|}{\bf Anomaly Detection (AD)} & \multicolumn{1}{c|}{\bf Explanation Discovery (ED)} & \multicolumn{1}{c|}{\bf Computational} \\
& \multicolumn{1}{c|}{\bf Functionality} & \multicolumn{1}{c|}{\bf Functionality} & \multicolumn{1}{c|}{\bf Performance} \\
\hline
\hline
{\bf Evaluation} & AD1: Anomaly Existence & ED1: Local Explanation & P1: AD Training Scalability \\
{\bf Criteria} & AD2: Range Detection & ED2: Global Explanation & P2: AD Inference Efficiency \\
& AD3: Early Detection & & P3: ED Efficiency \\
& AD4: Exactly-Once Detection & & \\
\hline
{\bf Evaluation} & Accuracy: \textit{Range-based Precision,} & Conciseness & Time, given  \\
{\bf Metrics} & \,\,\,\,\,\,\,\,\,\,\,\,\,\,\,\,\,\,\,\,\,\,\,\,\, \textit{Recall, F-Score, AUPRC} & Consistency: \textit{Stability} (ED1), \textit{Concordance} (ED2) & different \textit{Dimensionality}  \\
& & Accuracy: \textit{Point-based Precision, Recall, F-Score} & and \textit{Cardinality} factors \\
\hline
\end{tabular}
}
\vspace{-0.05in}
\caption{\small The \system\ evaluation methodology and benchmark design}
\vspace{-0.1in}
\label{tab:benchmark-design}
\end{table*}

\noindent
{\bf Disturbed Traces.} 
Disturbed traces 
were obtained by introducing anomalous events during an execution. Based on discussions with industry contacts from the Spark ecosystem, we came up with 6 types of anomalous events. When designing these, we considered that:
(i) they lead to a visible effect in the trace,
(ii) they do not lead to an instant crash of the application (since AD would be of little help in this case),
(iii) they can be tracked back to their root causes.
We briefly describe these 
anomalies below; please 
see Appendix~\ref{appendix:traces} for further details.

\minipb{Bursty Input (Type 1):}
To mimic input rate spikes, we ran a disruptive event generator (DEG) on the Data Senders to temporarily increase the input rate by a given factor for a duration of 15-30 minutes. We repeated this pattern multiple times during a given trace, creating a total of 29 instances of this anomaly type over 6 different traces. Please see Figure~\ref{fig:burstyinput} for an example.

\minipb{Bursty Input Until Crash (Type 2):}
This is a longer version of Type 1 anomalies, 
where we let the DEG period last forever,
crashing the executors due to lack of memory. When an executor crashes, Spark launches a replacement, but the sustained high rates 
keeps crashing the executors, until %
Spark eventually decides to kill the whole application. We injected this anomaly into 7 different traces. 

\minipb{Stalled Input (Type 3):}
This type of anomaly mimics failures of Spark data sources (e.g., Kafka or HDFS). To create it, we ran a DEG that set the input rates to 0 for about 15 minutes, and then periodically repeated this pattern every few hours, giving us a total of 16 anomaly instances across 4 different traces. 

\minipb{CPU Contention (Type 4):}
The YARN resource manager cannot prevent external programs from using the CPU cores that it has allocated to Spark processes, causing scheduling delays to build up due to CPU contention. We reproduced this anomaly using a DEG that ran Python programs to consume all CPU cores %
on a given Spark node. We created 26 such anomaly instances over 6 different traces. 

\minipb{Driver Failure (Type 5) and Executor Failure (Type 6):}
Hardware f\-aults or maintenance operations may cause a node to fail all of a sudden, making all processes (drivers and/or executors) located on that node  unreachable. Such processes must be restarted on another node, which causes delays. We created such anomalies by failing driver processes, where the number of processed records drops to 0 until the driver comes back up again in about 20 seconds. We also created anomalies by failing executor processes, which get restarted 10 seconds after the failure, but 
whose effects on metrics such as processing delay may continue longer. We created 9 driver failures and 10 executor failures over 11 different traces.

\noindent
\textbf{Ground Truth Table.}
For all of these 97 anomaly instances over 34 anomalous traces, we provide ground truth labels with the information shown in Table~\ref{tab:traces}. Such labels include both {\em root cause intervals} (RCIs) and their respective {\em extended effect intervals} (EEIs). RCIs typically correspond to the time period during which DEG programs are running, whereas the EEIs are the time periods that start immediately after an RCI and end when important system metrics return to normal values or the application is eventually pushed  to crash. 
The EEIs are manually determined using domain knowledge. Additional details can be found in Appendix~\ref{appendix:extended-effect}.

\ssection{Benchmark Design} 
\label{sec:benchmark}

In this section, we present the evaluation methodology we designed to benchmark anomaly detection (AD) and explanation discovery (ED) algorithms based on the curated, high-dimensional time series dataset described in the previous section. As summarized in Table~\ref{tab:benchmark-design}, \system\ is designed to evaluate AD and ED algorithms in two orthogonal aspects, \textit{functionality} and \textit{computational performance}, using well-defined   metrics. In terms of functionality, the evaluation criteria capture that an AD/ED algorithm is exposed to increasingly more challenging requirements as the functionality level is raised from one to the next. In terms of computational performance, \system\ provides three complementary criteria that can be evaluated by varying dimensionality and size of the dataset. 

\vspace{-0.1cm}
\ssubsection{Anomaly Detection (AD) Functionality}

First and foremost, we designed \system\ targeting {\em semi-supervised AD techniques} (i.e., trained only with normal data, possibly with occasional noise, and then tested against anomalous data) for {\em range-based anomalies} (i.e., contextual and collective anomalies occurring over a time interval instead of only at a single time point) over {\em high-dimensional} (i.e., multivariate with 1000s of dimensions) time series. This decision is informed by our observation of this being the most common and inclusive usage scenario in practice.

\noindent
\textbf{Evaluation Criteria.}
We identified four key criteria for evaluating AD functionality, listed below from basic towards advanced, where a higher AD level includes the requirements of all preceding levels: 

\noindent
\underline{AD1 ({\em Anomaly Existence}):} The first expectation is to flag the existence of an anomaly somewhere within the {\em anomaly interval}
(i.e., the root cause interval (RCI) + the extended effect interval (EEI)).

\noindent
\underline{AD2 ({\em Range Detection}):} The next expectation is to report not only the existence, but also the precise time range of an anomaly. The wider a range of an anomaly that an AD method can detect, the better its understanding of the underlying real-world phenomena.

\noindent
\underline{AD3 ({\em Early Detection}):} The third expectation is to minimize the {\em detection latency}, i.e., the difference between the time an anomaly is first flagged and the start time of the corresponding RCI.

\noindent
\underline{AD4 ({\em Exactly-Once Detection}):} The last expectation is to report each anomaly instance exactly once. Duplicate detections are undesirable,
because they may not only redundantly cause repeated alerts for a single anomalous event, but also confusion if those alerts are for the same anomaly event or not.

\begin{figure}[t]
\begin{center}
\includegraphics[width=\columnwidth]{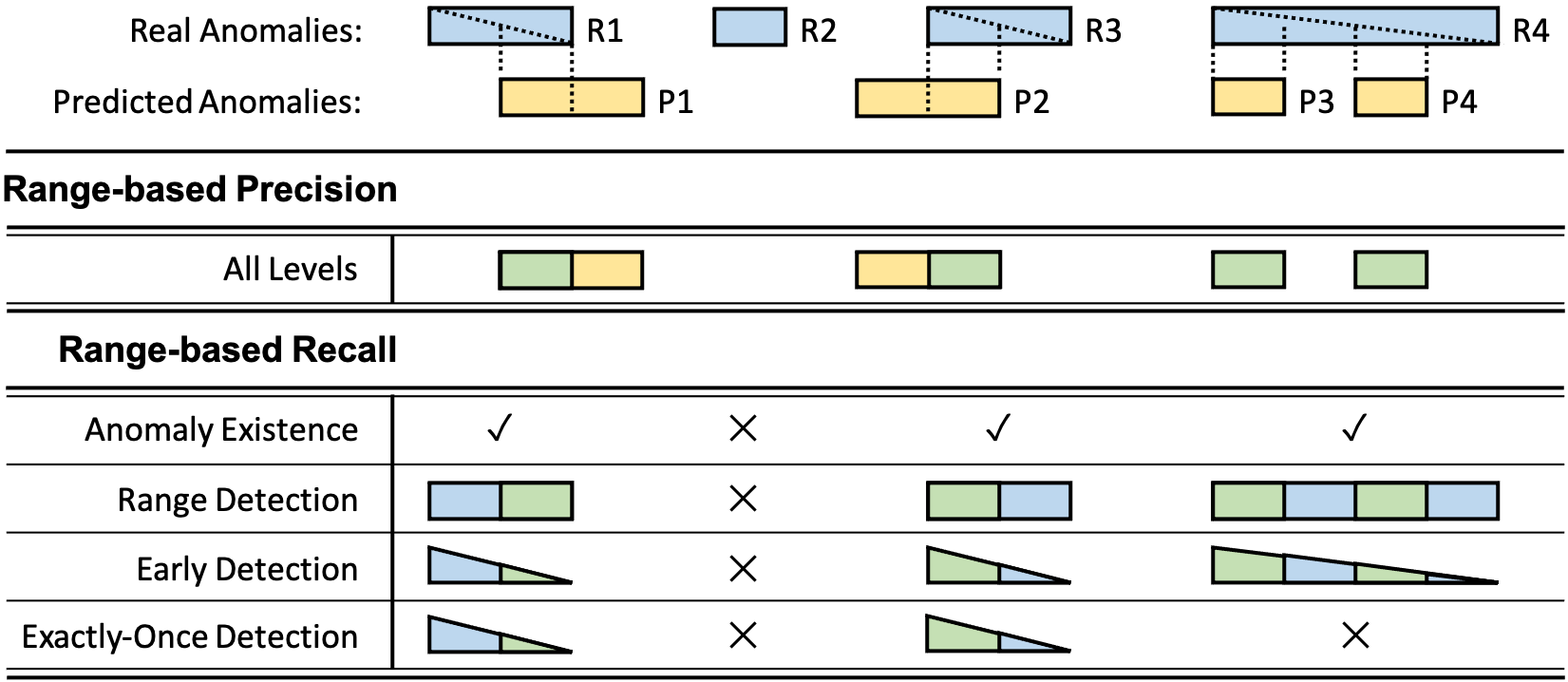}
\vspace{-0.3in}
\caption{\small Range-based precision and recall at  AD levels 1-4. Precision evaluates prediction quality (green out of yellow for each Pi).  Recall evaluates anomaly coverage (green out of blue for each Ri).}
\vspace{-0.3in}
\label{fig:AD-levels}
\end{center}
\end{figure}

\noindent
\textbf{Evaluation Metrics.}
To assess how well an AD algorithm can meet these four functionality levels, we use the customizable accuracy evaluation framework for time series~\cite{TatbulLZAG18}.
This framework extends the classical precision/recall from point-based data to range-based data, by introducing a set of tunable parameters. By setting the values of these parameters in a particular way and applying the resulting precision/recall formulas to the output of an AD algorithm, one can assess how well that output measures up to the quality expectations represented by those parameter settings. We leverage this as a mathematical tool to quantify how well an AD algorithm meets AD1-AD4. Furthermore, we chose to do this in a way that every level AD$_i$ builds on and adds to the requirements of the previous level AD$_{i-1}$. This monotonic design ensures that the AD functionality score that an algorithm gets (Precision, Recall, or other metrics obtained by combining them, e.g., F-Score or Area Under the Precision/Recall Curve (AUPRC)) is always ordered as: $score(AD1)$ $\geq score(AD2) \geq score(AD3) \geq score(AD4)$, which facilitates evaluating and interpreting results in a systematic way. In \system, we preferred this design over the alternative of treating each AD level as orthogonal to enable users to develop/perfect their models for tasks that become increasingly more challenging.

Figure~\ref{fig:AD-levels} provides a simple example to illustrate how range-based precision and recall are computed for different AD levels. Given real anomaly ranges R1..R4 and predicted anomaly ranges P1..P4 produced by an AD algorithm, we first compute precision/recall for each range and then average them for overall precision/recall. Intuitively, precision focuses on the size of TP ranges (colored green) relative to TP+FP ranges (colored yellow), and recall focuses on the size of TP ranges relative to TP+FN ranges (colored blue). For AD1, Recall(Ri) is 1 if Ri is flagged, 0 otherwise. For AD2, Recall(Ri) is proportional to the relative size of the TP range. For AD3, Recall(Ri) is further weighted by position of the TP range relative to the start of Ri. Finally, at AD4, Recall(Ri) degrades to 0 for any Ri that is not flagged exactly once. Precision(Pi) is computed in an analogous way, except that AD levels about anomaly coverage quality (AD1 and AD3) are not relevant to it; rather, the main focus is on the size and number of the real anomaly ranges that are successfully predicted. In our simple example, it turns out that all AD levels for Precision consider the same Pi subranges.

To achieve precision and recall at different AD levels, we set the tunable parameters of the range-based precision/recall framework with necessary modifications, presented in Appendix~\ref{appendix:ad-functionality}.
Further note that both the semi-supervised AD algorithms we investigate and the precision/recall for time series model we use to assess them focus on binary classification (normal vs. anomalous ranges). On the other hand, our dataset is inherently a multi-class one (normal ranges vs. six types of anomalous ranges). This raises a question about how to evaluate binary predictions under multi-class labels. We take a holistic approach, and evaluate the AD prediction results both globally and grouped by type, whenever this is reasonable and provides useful insights. For example, even though a binary predictor is not able to detect different anomaly types, we can still measure its resulting coverage (i.e., recall) for each type. However, type-wise measurement is not entirely meaningful for
precision, since false positives (FPs) are essentially typeless.

Our benchmark also includes a set of four learning settings {\em LS1-LS4}, ranging from simple to more complex yet realistic ones, detailed in Appendix~\ref{appendix:ad-functionality}.

\ssubsection{Explanation Discovery (ED) Functionality} \label{sec:ed-functionality}

Once an anomalous instance is flagged by an AD method, the next desirable functionality is to find the  best explanation for the anomaly detected, or  more precisely, a human-readable formula offering useful information about what has led to the anomaly. 

There have been many ED methods in recent work (see \S \ref{sec:related}). These differ in the form of ``explanation'' provided: some return a logical formula as an explanation~\cite{ZDM2017,macrobase:sigmod17,Ribeiro0G18}, others return a decision tree~\cite{WuHPZ2018}, and some others return a numerical score for each feature  such as the coefficient in linear regression~\cite{Ribeiro0G16} or the SHAP score~\cite{LundbergL17}.
\system\ does not pose any restrictions on the form of explanation used. Instead, it takes an abstract view of explanations.
Formally, we model each trace in the test dataset as a multi-dimensional time series, $[\msbold{x}_1 \ldots \msbold{x}_t \ldots \msbold{x}_n]^T$, where each data item includes $m$ features, $\msbold{x}_t = (x_{t1}, \ldots, x_{tm})$.  A detected anomaly is a subsequence of the time series that starts at timestamp $t$ and has duration $w$, $X_{t,w} = [\msbold{x}_t \ldots \msbold{x}_{t+w}]^T$. If an AD method can provide only point-based detection, then $w$ is set to 0. 
We denote the explanation generated for the anomaly $X_{t,w}$ as $F_{t,w}$ and treat it as a function of the features, $\mathbf{A} = (a_1, \ldots, a_m)$, from the data:
$$F_{t,w} (a_1, \ldots, a_m) \models X_{t,w}$$ 
where $\models$ means that $F_{t,w}$  ``explains'' the anomaly $X_{t,w}$. 
In addition, we define an extraction function, $G_{\mathbf{A}}$ over $F_{t,w}$, that returns the set of features  used in the explanation (e.g., appearing in a logical formula or having non-zero coefficients in a regression model):
$$G_\mathbf{A} \left(F_{t,w} (a_1, \ldots, a_m)  \right) = \mathbf{A}_{t,w} \subseteq \mathbf{A} $$ 
Finally, we define the size of $F_{t,w}$  as the size of its feature set $\mathbf{A}_{t,w}$:
$$|F_{t,w} (a_1, \ldots, a_m)| = |\mathbf{A}_{t,w}|$$

\noindent
\textbf{Evaluation Criteria: Subject of Explanation.}
The key distinction that \system\ makes is whether an ED method is attempting to explain a single anomaly ({\em local}) or a broad set of anomalies ({\em global}).  

\noindent
\underline{ED1: \textit{Local Explanation}:}
This corresponds to explaining one anomaly instance, offering a compact yet meaningful piece of information to help the  user  understand this particular instance. As mentioned by LIME~\cite{Ribeiro0G16}, the explanation should be locally faithful. In our context, it means that the same explanation can hold over immediate ``neighbors'', which are anomaly instances of the same application and same anomalous type, and around the same time period.
 
\noindent
\underline{ED2: \textit{Global Explanation}:} Alternatively, an ED method may attempt to explain a (potentially large) set of anomalies, called a global explanation. 
In general, it is not possible to find an identical succinct explanation for many different instances. Hence, a global explanation is usually composed of a set of  explanations; e.g., LIME~\cite{Ribeiro0G16} chooses the most representative $k$ instances to explain a model. In our benchmark, it makes most sense to construct a global model for a set of anomalies of the same type, but potentially from different applications or different  runs of the same application. This helps us understand for ``semantically similar" anomalies, whether an ED method can return explanations that are consistent, 
or even of predictive power of similar anomalies that arise in the future.

\noindent
\textbf{Evaluation Metrics.} 
\system\ evaluates both local and global explanations for three desired properties: 

\noindent
\underline{1. {\em Conciseness}:} This corresponds to the number of features used in the explanation. Following the Occam's razor principle, humans favor smaller, and thus simpler explanations. 
As different ED methods return explanations of different forms, our benchmark counts the number of  features used in the explanation as its conciseness measure. In the ED1 case, that is $|F_{t,w}|$ defined above. In the ED2 case, a global explanation includes  a set of explanations, and its conciseness measure is the average of the size of each explanation.

\noindent
\underline{2. {\em Consistency}:} Anomalies of the same type occurring in a similar context %
should have consistent explanations. 
We customize this notion for ED1 and ED2, respectively. %
In both cases, we care only about  the set of features employed in the explanation, without considering the numerical or categorical values used.

{\em Stability} (ED1) is the customized consistency measure for ED1. It means that the anomalies occurring in a similar context (e.g., for the same application, same run, and same time period) should have similar explanations, subject to a small perturbation of the data. 
Formally, we introduce a subsampling procedure over an anomaly $X_{t,w}$, which generates a set of samples, $\{ X_{t,w}^{(i)} \}$. We denote the corresponding explanations generated for them  as $\{ F_{t,w}^{(i)} \}$. 
The extraction function for a set of explanations is defined to be the duplicate-preserving union (like \textsc{Union All} in SQL) of the extraction function of each respective explanation: 
$$G_\mathbf{A} \left( \big\{F_{t,w}^{(i)} (a_1, \ldots, a_m) \big\} \right)  = \bigcup_{i} G_\mathbf{A} \left(F_{t,w}^{(i)}  \right) = \bigcup_{i} \mathbf{A}_{t,w}^{(i)} =   \mathbf{A}_{t,w}^{\cup} $$ 
Finally, for each feature $a_j \in \mathbf{A}_{t,w}^{\cup} $, we count its frequency in this feature set and normalize it by the total size of the feature set. The consistency measure is then defined as the entropy of the set of normalized frequencies of such features, $a_j$, $j=1, 2, \ldots$: 
$$ H\left(A_{t,w}^{\cup} \right)  = - \sum_j p(a_j) \log_2 p(a_j), \,\,\,\,\,\, a_j \in\mathbf{A}_{t,w}^{\cup} $$
$$ p(a_j) = \mathbbm{1}_{\mathbf{A}_{t,w}^{\cup}} \left(a_j \right) \,\, / \,\, \bigr|  \mathbf{A}_{t,w}^{\cup} \bigl| \,\,\,\,\,\,\,\,\,\,\,\,\,\,\,\,\,\,\,\,\,$$

Where $\mathbbm{1}_{\mathbf{A}_{t,w}^{\cup}} \left(a_j \right)$ is here an indicator function that counts the occurrences of a feature in a multiset. 
For capturing consistency, our choice of entropy is motivated by information theory that a set of explanations that lack consistency will require using more bits to encode, hence a larger entropy value. In the ideal case, all  explanations,  $\{ F_{t,w}^{(i)} \}$, are identical, and its entropy takes the minimum value 0  if the size of the explanation is 1 (denoted as $H_1$),  the value 1 if the size is 2 ($H_2$), or the value 1.58 if the size is 3 ($H_3$). 

{\em Concordance} (ED2) is the customized consistency measure for ED2. Here, it means that the anomalies of the same type are expected to have consistent explanations, subject to larger amounts of deviation in data due to different time periods in the same run of a Spark application, different runs of the application, or even different Spark applications.   
Formally, we are given a set of anomalies, $\{ X_{t_i,w_i} \}$. Denote their corresponding explanations as $\{ F_{t_i,w_i} \}$. The consistency measure of this set of explanations is computed similarly to 
ED1, except that we are replacing the subsampled anomalies, $\{ X_{t,w}^{(i)} \}$, with the given set of anomalies, $\{ X_{t_i,w_i} \}$.

However, one may notice that the conciseness measure also has an impact on consistency. To factor out this impact, we further define {\em Normalized Consistency} as $\frac{2^{Consistency}}{Conciseness}$, which  captures the variability of the explanations conditioned on their average size.

\noindent
\underline{3. {\em Accuracy}:} The last property, which is also the hardest to achieve, is to view an explanation of an anomaly as a predictive model, apply it to  other similar  instances (defined above for ED1 and ED2, respectively), and then evaluate accuracy 
of such predictions. 

Note that not all explanations can serve as a predictive model. Only those that are a function mapping a given data item to 0/1, $ F_{t,w}: \msbold x_t\in \mathbb{R}^m \rightarrow \{0, 1\}  $, can offer predictive power over test data.
For example, a logical formula~\cite{ZDM2017,macrobase:sigmod17,Ribeiro0G18} or a decision tree~\cite{WuHPZ2018}  can be used to run prediction on new data items, but feature importance scores or SHAP scores~\cite{LundbergL17} cannot. 
Even with those ED methods that return a predictive explanation, it is only a point-based predictive model. 
The literature largely lacks ED methods that can return  explanations that characterize a temporal pattern. 
For this reason, \system\ evaluates the accuracy of such explanations using point-based precision and recall. 

In the case of ED1, we are given a particular anomaly $X_{t,w}$. To measure the accuracy of an ED method, we subsample from $X_{t,w}$, yielding a sample, $X_{t,w}^{(i)}$. We run the ED method to generate an explanation, $F_{t,w}^{(i)}$. Then we run $F_{t,w}^{(i)}$ as a predictive model over a test dataset that includes the remainder of the anomalous  data, $X_{t,w} - X_{t,w}^{(i)}$, as well as some normal data that immediately proceeds or follows $X_{t,w}$. For each test point, we obtain a 0/1 prediction and compare it to the ground truth.
We repeat this procedure for all test points to compute the final precision, recall, and F-score. %
 
In ED2, we are given a set of anomalies $X_{t_i,w_i}$. We randomly split this set into a training set and a test set. We can run a suitable ED method to generate a global explanation from the training set, and then use it as a predictive model over the test set. For each anomaly in the test set, we compare the point-wise prediction against the ground truth and compute  precision, recall, F-score, similar to ED1. 

\ssubsection{Computational Performance}

\system\ can also be used to evaluate computational performance.

\noindent
{\bf Evaluation Criteria and Metrics.}
ML algorithm performance is typically measured in terms of the total time it takes for model training as well as for using that model for making predictions. For AD, we define P1 and P2 to evaluate training and inference performance, respectively. For ED, the time to discover each explanation, P3, is our third performance metric.

\noindent
{\bf Experimental Parameters.}
\system\ offers scalability tests by varying the following two data-related parameters:

\noindent
\underline{\em Dimensionality $M$:}
Our dataset consists of high-dimensional time series data. The 2,283 metrics (features) may be correlated and contain a lot of null values, which are representative of real-world datasets. The benchmark leaves it to each user algorithm as how it copes with the high dimensionality. The relevant techniques may include dimensionality reduction using linear transformation (e.g., PCA), 
or feature selection by leveraging the correlation structure in the data. Such choices are left to the discretion of each user algorithm, and \system\ reports on the resulting dimensionality $M$ used in AD and ED tasks.

\noindent
\underline{\em Cardinality Factor $\alpha$:}
Besides high dimensionality, our dataset also has high cardinality, $N =$  2,335,781 data items, which is significantly higher than the existing Numenta Anomaly Benchmark~\cite{numenta:icmla15}. If the training time of an algorithm is too long, 
a user algorithm  can choose to reduce the cardinality via resampling, i.e., by taking average of the data items in each $l$-second interval, which amounts to a cardinality factor $\alpha=1/l$ and reduced data size of $\alpha N$.

\ssubsection{Broader Applicability} 
\label{sec:applicability}

The \system\ benchmark is more broadly applicable beyond our particular dataset (see Appendices \ref{appendix:ad-applicability} and \ref{appendix:ed-applicability} for more details):

\noindent
\textbf{AD Benchmark.} 
The \system\ benchmark considers all techniques that  1) train a model for data normality via learning from undisturbed traces, 2) assign outlier scores to new test records, and 3) derive binary predictions from these scores using a threshold.
When meeting the above conditions, our AD metrics  can be used with any labeled time series anomaly test datasets similar to ours. The four AD levels can be directly usable if labels are available as ranges (like for real or synthetic datasets from the discord discovery literature, also used in~\cite{seq2graph}, \cite{norm}), while one would need to set our evaluation parameters to  classical precision and recall if labels are available only as points (like for classification-oriented datasets tuned to data points of imbalanced classes, some of which are used in~\cite{Tran2015}).
Applying such metrics also helps assess the performance and efficiency of any time series AD technique capable of assigning outlier scores and binary predictions to each record of a test sequence. 

\noindent
\textbf{ED Benchmark.} While a user study may be the best way to evaluate the usefulness of explanations, it is not always available and may come at a high cost. Therefore, our benchmark aims to provide \textit{automated} evaluation of ED methods based on intuitive metrics, namely, \textit{conciseness}, \textit{consistency}, and \textit{accuracy}, as well as their various variants in the ED1 and ED2 settings. 
As further detailed in Appendix~\ref{appendix:ed-applicability}, our metrics cover many of the metrics used in prior ED works, except for specific metrics that depend on a particular model or algorithm, which \system\ deliberately avoids as a general benchmark, or require ground truth features or visual inspection by domain experts, which are not always available in complex domains. 
As the result of sharing metrics with existing works, the ED metrics of \system\ can be applied to the datasets used in~\cite{ZDM2017,macrobase:sigmod17}, as well as those in~\cite{LakkarajuBL16, WuHPZ2018,Ribeiro0G18} for the sake of evaluating the explanation for a classification result, and those 
in~\cite{Ribeiro0G16,LundbergL17} with sufficient preprocessing on the text or image data used.

\vspace{0.1in}
\ssection{A Full Pipeline for Explainable AD}
\label{sec:pipeline}

Besides a curated dataset and an evaluation methodology, \system\ also provides a full pipeline for explainable AD on high-dimensional time series data. Our pipeline is characterized as follows:
(i)~It consists of the typical steps in a deep ML pipeline, ranging from data partitioning, feature engineering, dimensionality reduction, to AD and ED.
(ii)~It implements a variety of AD and ED functionalities and evaluation modules that score them based on the metrics of the benchmark (see \S \ref{sec:benchmark}).
(iii)~It provides an open, modular architecture that allows different methods to be added and combined through the pipeline.
Figure~\ref{fig:pipeline} provides an overview.

\minipa{1. Data Partitioning.}
The first phase takes as input the 93 raw traces, described in \S \ref{sec:anomalies}. It performs simple data cleaning, e.g., replacing missing data with a default value. It then performs data  partitioning of the 93 traces. In the default  setting,  we take all undisturbed traces as training data, 
$D_{\text{train}}$, and all disturbed traces as test data, 
$D_{\text{test}}$.
Other implementation choices are left to Appendix~\ref{appendix:pipeline}. 

\minipa{2. Data Transformation.}
As ML algorithms require data transformations  to perform well, our pipeline offers the following steps:

\noindent
\underline{(i) \textit{Resampling} (optional):} %
For the multivariate time series in each trace, the user can choose to resample, by taking the average of data points in each $l$-second interval. 
This step reduces the {\em cardinality factor}, $\alpha = 1/l$, of the time series data, if the training time turns out to be too long for some ML algorithms.

\noindent
\underline{(ii) \textit{Dimensionality reduction}:} Since our dataset includes $M=2,283$ raw features, 
such high dimensionality may affect both model accuracy, known as the ``curse of dimensionality'', and training time. 
To reduce dimensionality, our pipeline offers a PCA-based method (with a parameter that controls different coverage of the data variance and the resulting feature set size), as well as a manually curated feature set with 19 features selected using domain knowledge.

\noindent
\underline{(iii) \textit{Rescaling}:} Most ML algorithms require the features to be scaled into a range, e.g., between [0, 1], to better align features whose raw values may differ by orders of magnitude. 
A unique issue in our problem  is that each test trace may represent a new context, e.g., a combination of input rate and concurrency  not seen in training data. As a result, rescaling has to take into account this new context. To simplify this setting, we provide the option of performing rescaling per trace, as well as a customized scaling method that rescales test data dynamically as we run an AD model over the data.

\minipa{3. AD Modeling.}
The next phase takes the transformed training data and builds an AD model. Most AD methods build a model that describes the normal behavior in the data, called a ``normality model'', such that any future (test) data that deviates significantly from it will be flagged as an anomaly. 
Our pipeline offers an open architecture to embrace any AD method that builds such a normality model to detect anomalies.
In this paper, we focus on recent DL-based AD methods \cite{MVSA, XuCZLBLLZPFCWQ18, SchleglSWSL17}, to explore their potential for handling  the complexity of our dataset (high-dimensional, with noise), anomaly patterns (a variety of contextual and collective anomalies), and learning settings (noisy semi-supervised AD modeling).

\noindent
\underline{(i) {\em Normality modeling}:} 
The first step is to train a normality model based on a DL method of choice.
Most DL methods take input data of fixed window size $s$. Given each of our traces, we create sliding windows of size $s$ and slide 1, and feed them as input to the model. 
Different DL methods model the data in the window by either trying to forecast the data point  following the window (forecasting-based, e.g., LSTM~\cite{BontempsCML16}) 
or reconstructing the window via a succinct internal representation (reconstruction-based, e.g., Autoencoder~\cite{HS,XuCZLBLLZPFCWQ18} or GANs~\cite{SchleglSWSL17}). 
To train each specific model, we divide the transformed $D_{\text{train}}$ set into internal training ($D^0_{\text{train}}$), validation ($D^1_{\text{train}}$), and test ($D^2_{\text{train}}$) sets. The DL model is trained on $D^0_{\text{train}}$, with early stopping applied based on the model performance on $D^1_{\text{train}}$. Hyperparameter tuning is performed by choosing a configuration that maximizes model performance on $D^2_{\text{train}}$.

\noindent
\underline{(ii) {\em Outlier score derivation}:} We next build an initial AD model, $g: \msbold x\in \mathbb{R}^m \rightarrow \mathbb{R}$, which maps each data point to an outlier score.
For forecasting models, we compute the difference, $d$, between the forecast and true values of each data point, and derive the outlier score $v$ based on $d$; the higher the $d$ value, the higher the $v$ score.  
For reconstruction-based models, we treat the reconstruction error of each window as the $v$ score for that window, and then derive the $v$ score of each data point by averaging the scores of its enclosed sliding windows. 
Our pipeline implements the LSTM~\cite{BontempsCML16}; Autoencoder (AE)~\cite{HS}; and BiGAN~\cite{SchleglSWSL17} for AD. 
Details of these models and the necessary modifications we made to suit our dataset can be found in Appendix~\ref{appendix:ad-methods}.

\noindent
\underline{(iii) {\em Threshold selection}:} 
The last step aims to find a threshold on the outlier score to return a 0/1 prediction. 
It returns a final AD model, $f: {\msbold x}\in \mathbb{R}^m \rightarrow \{0, 1\}$, mapping each data point to 0/1. 
\system\ does not offer labeled data for threshold selection. Hence, we provide unsupervised threshold selection fit on $D^2_{\text{train}}$. Among the methods listed in a recent survey~\cite{YangRF19}, we choose three most used automatic techniques: SD, MAD, and IQR, with the possibility of repeating them multiple times to filter large outlier scores.

\minipa{4. AD Inference.} Once the AD model is built, the next phase of the pipeline runs the AD model over each test trace to detect anomalies. In the context of range-based AD, predicted anomalies for a test trace are defined as sequences of positive predictions within that trace, denoted as $X_{t,w}$, which starts at $t$ and has duration $w$. 

\minipa{5. AD Evaluation.}
The last AD phase evaluates the AD model for a given set of requirements. We  evaluate both a model's ability to separate normal from anomalous data in the outlier score space and its final AD ability based on threshold selection.
The separation ability ($g$) is assessed at the trace, application, and global levels. Global separation is reported as the AUPRC computed on all test data, while the application/trace-level separation is reported by computing an AUPRC for each application/trace, and averaging the results. %
The detection ability ($f$) is assessed by reporting its range-based precision, recall, and F-score, with parameters specified by the AD functionality. Recall is also reported by anomaly type.

\minipa{6. ED Execution.} 
For each test trace, AD inference  reports a set of anomalies, and for each reported anomaly, the ED module returns an explanation for it. Our pipeline supports two families of ED methods.
(i) {\em Model-free ED methods} do not require the access to an ML model. Instead, they only require the anomalous instance, $X_{t,w}$, and a reference dataset, to generate an explanation. Examples include EXstream~\cite{ZDM2017} and MacroBase~\cite{macrobase:sigmod17} (\S \ref{sec:related}). Our implementation sets the reference dataset as the subset of data that immediately proceeds the detected anomaly, denoted by $X_{t,-w'}$, and was classified as normal. Then the pair of datasets, $(X_{t,w}$, $ X_{t,-w'})$, are provided to the ED method to generate an explanation, $F_{t,w}$.
(ii) {\em Model-dependent ED methods} take not only the anomalous instance, $X_{t,w}$, but also an AD model, $f: \msbold{x}\in \mathbb{R}^m \rightarrow \{0,1\}$. 
Examples include LIME~\cite{Ribeiro0G16}, Anchors~\cite{Ribeiro0G18}, and SHAP~\cite{LundbergL17} (\S \ref{sec:related}).
In our implementation, we provide the AD model used in inference to the ED method. 

\minipa{7. ED Evaluation.} After processing each test trace, we obtain a set of anomalies with their corresponding explanations. We then collect  the explanations from all the test traces to run the final ED evaluation and compute 
conciseness, consistency, accuracy, and time metrics. 
Further details are given in Appendix~\ref{appendix:pipeline}.

\ssection{Experimental Study} 
\label{sec:experiments}

In this section, we apply our benchmark to a select set of AD and ED methods.
While a comprehensive comparison of related AD and ED methods is beyond the scope of this paper, analyzing the select methods allows us to demonstrate the value of our dataset and benchmark. Our analyses include the strengths and limitations of these AD and ED methods, challenges posed by our dataset and evaluation criteria, and some potential directions of future research. 

\ssubsection{Experimental Setup}

In our experimental setup, we integrated into our pipeline 
three DL-based AD methods: 
LSTM~\cite{BontempsCML16}, AE~\cite{HS}, and BiGAN \cite{SchleglSWSL17}. 
We also integrated  three recent ED methods:
EXstream~\cite{ZDM2017} and MacroBase~\cite{macrobase:sigmod17}, from the DB community for outlier explanation in data streams, and LIME~\cite{Ribeiro0G16}, an influential method from the ML community. 
Details regarding these methods can be found in 
Appendices~\ref{appendix:ad-methods} and \ref{appendix:ed-methods}.

Besides the methods, our pipeline also needs to be configured with the following options:
(a)~{\bf Data Size and Feature Set (FS)}: Since some of the DL models (e.g., GANs) could not complete training on our cluster using the full dataset, we reduced the data size by setting the cardinality factor, $\alpha = 1/15$. We also used a reduced feature set of 19 features, $m=19$, produced by either manual selection based on domain knowledge, denoted as FS$_{\text{custom}}$, or by PCA with the same number (19) of features, denoted as FS$_{\text{pca}}$. 
For each given training set, we allowed each DL algorithm to train for 1.5 days (including hyperparameter tuning) to obtain an AD model. 
(b)~{\bf Level of AD Evaluation (AD1-4)}, as described  in Table~\ref{tab:benchmark-design}, with a default setting of AD2 (range detection). 

\ssubsection{AD Evaluation Results and Discussion}

\begin{figure*}[t]
	\centering
	\begin{tabular}{lc}
		\hspace{-0.2in}
		\subfigure[\small{Trace-wise separation: T2 trace of Application 2}]
		{\label{fig:lm4-ae-trace-separation-app2-t2}\includegraphics[width=0.48\textwidth,height=2.2cm]{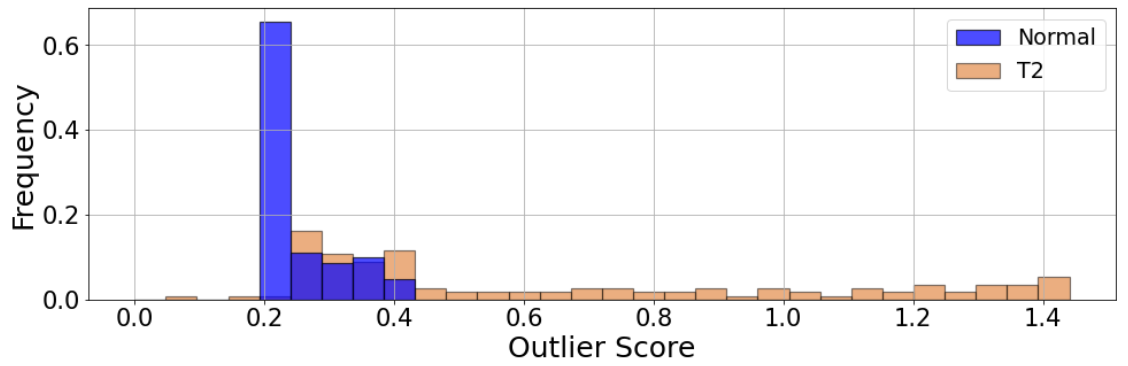}}
		&
		\subfigure[\small{App-level separation: all disturbed traces of Application 2}]
		{\label{fig:lm4-ae-app-separation-app2}\includegraphics[width=0.48\textwidth,height=2.2cm]{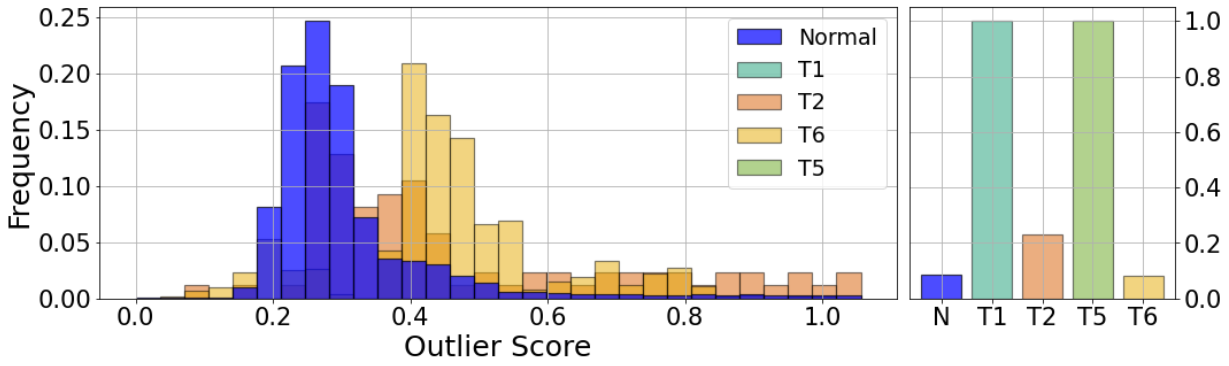}} \\
		\subfigure[\small{Global-level separation: all disturbed traces with ``best'' AD2 threshold}]
		{\label{fig:lm4-ae-ad2-global-separation}\includegraphics[width=0.48\textwidth,height=2.2cm]{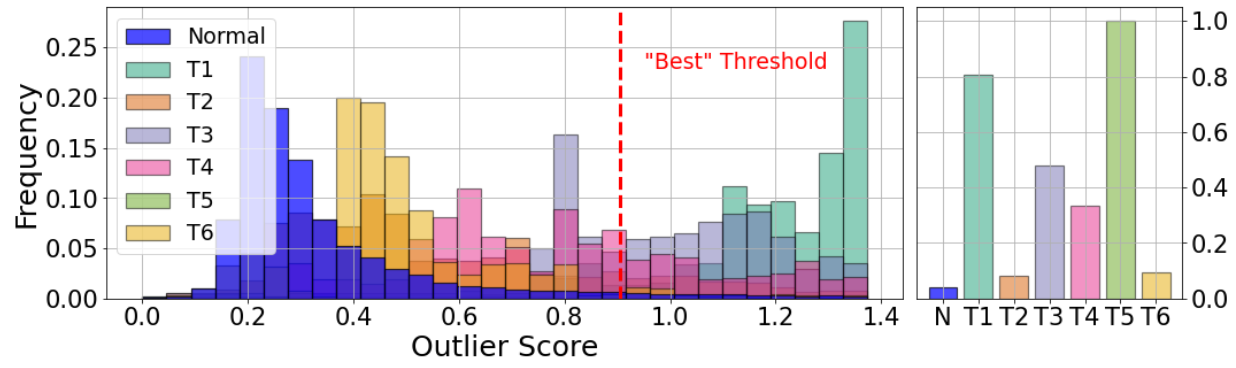}}
		&
		\subfigure[\small{Modeling test samples with ``best'' AD2 threshold}]
		{\label{fig:lm4-ae-ad2-m-test-scores}\includegraphics[width=0.48\textwidth,height=2.2cm]{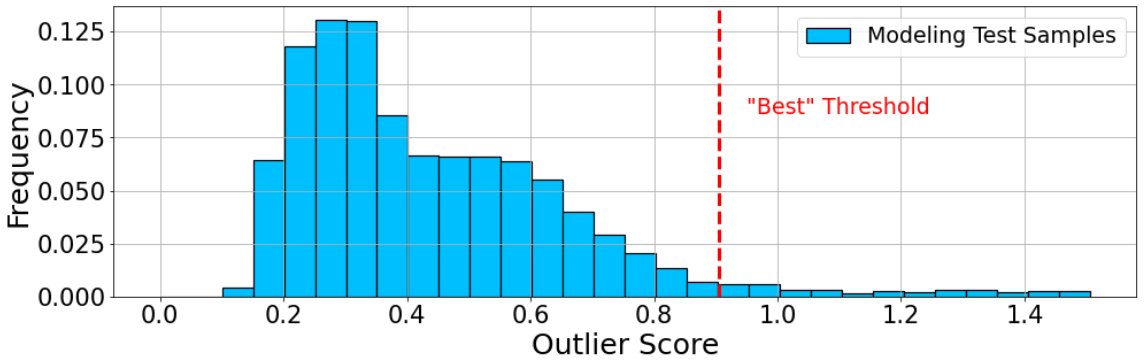}} \\
		\subfigure[\small{Record-wise outlier scores on a T1 trace of Application 2}]
		{\label{fig:lm4-ae-app2-bursty}\includegraphics[width=0.48\textwidth,height=1.8cm]{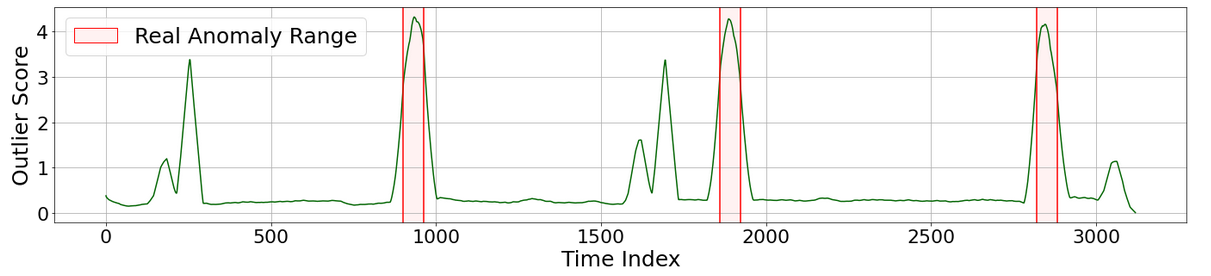}}
		&
		\subfigure[\small{Record-wise outlier scores} on a T2 trace of Application 2]
		{\label{fig:lm4-ae-app2-bursty-crash}\includegraphics[width=0.48\textwidth,height=1.8cm]{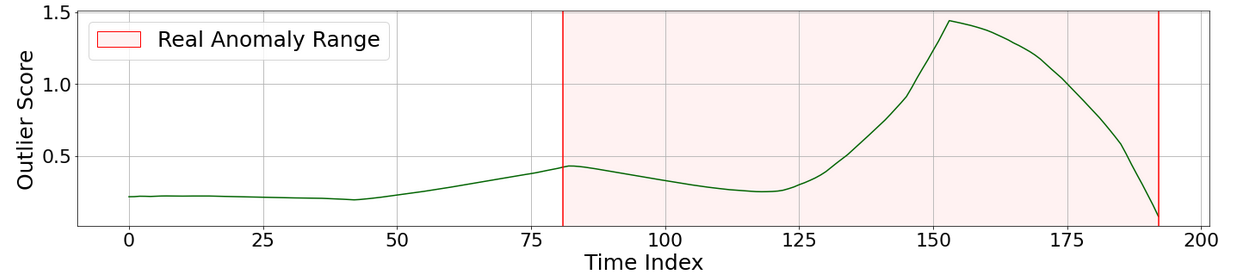}}
	\end{tabular}
\vspace{-0.2in}
\caption{\small Outlier score distributions and record-wise outlier scores using the AE method (FS$_\text{custom}$)}
\label{fig:lm4-ae-ad2-scores-dists}
\vspace{-0.1in}
\end{figure*}

We begin by applying the  LSTM~\cite{BontempsCML16}, AE~\cite{HS}, and BiGAN~\cite{SchleglSWSL17} methods on our benchmark dataset and report on the AD  metrics. 

\minipa{Experiment 1 (FS$_{\text{custom}}$, AD2).} 
The first experiment compares the three AD methods under a default setting, (FS$_{\text{custom}}$, AD2). 
Here, we focus on the model's ability to separate anomalous data from normal data, via the analysis of trace-level, application-level, and global AUPRC results summarized in Table~\ref{tab:custom-ls4-ad2-separation}.

\begin{table}[t]
    \centering
    \footnotesize
    \begin{tabular}{|c|c|c|cccccc|}
    	\hline
    	 Sep Lvl & Method & Ave & \multicolumn{6}{c|}{AUPRC for Anomaly Types T1$\rightarrow$T6}\\ \hline \hline
   		\multirow{3}{*}{\textbf{Trace}} & 
   		\textbf{LSTM} & 0.60 & 0.69 & 0.81 & 0.43 & 0.45 & 0.77 & 0.44 \\
   		& \textbf{AE} & {\bf 0.73} & 0.83 & 0.81 & 0.64 & 0.76 & 0.89 & 0.44 \\
   		& \textbf{BiGAN} & 0.61 & 0.91 & 0.76 & 0.15 & 0.70 & 0.64 & 0.51 \\ \hline \hline
   		\multirow{3}{*}{\textbf{App}} & 
   		\textbf{LSTM} & 0.47 & 0.57 & 0.37 & 0.56 & 0.38 & 0.60 & 0.35 \\
   		& \textbf{AE} & {\bf 0.57} & 0.65 & 0.40 & 0.63 & 0.55 & 0.79 & 0.43 \\
   		& \textbf{BiGAN} & 0.52 & 0.81 & 0.36 & 0.25 & 0.54 & 0.69 & 0.48 \\ \hline \hline
   		\multirow{3}{*}{\textbf{Global}} & 
   		\textbf{LSTM} & 0.41 & 0.56 & 0.32 & 0.53 & 0.25 & 0.53 & 0.27 \\
   		& \textbf{AE} & {\bf 0.50} & 0.60 & 0.36 & 0.54 & 0.47 & 0.68 & 0.37 \\
   		& \textbf{BiGAN} & 0.49 & 0.68 & 0.32 & 0.39 & 0.52 & 0.65 & 0.39 \\ \hline
    \end{tabular}
    \caption{\small Separation abilities of  AD methods (FS$_{\text{custom}}$, AD2)} \label{tab:custom-ls4-ad2-separation}
    \vspace{-0.3in}
\end{table}

(1) \textit{Trace-level Separation}: We first consider trace-level separation. All three methods achieved decent  AUPRC scores for most (or a subset) of anomaly types, with AE achieving the highest score of 0.73. Figure~\ref{fig:lm4-ae-trace-separation-app2-t2} shows the distribution of outlier scores assigned by the AE method to the records in the T2 trace of Application 2. 
In this example, the normal records are separated from anomalous ones for most the data. This shows that our data indeed carries useful signals that can be picked up by the AD method, which allows the AD method to perform better than naive classifiers that randomly assign a normal or abnormal label, or assign each instance to the majority (normal) class -- we refer to this remark as \textbf{R1}. 

(2) \textit{Application-level and Global Separation}:
Moving from the trace- to application- to global level separation, the AUPRC scores gradually decrease. This is because the separation of normal from anomalous instances in outlier score becomes increasingly harder as we broaden the contexts in which data is generated. Figure~\ref{fig:lm4-ae-app-separation-app2} shows the distributions of outlier scores assigned by the AE method to all disturbed traces of Application 2.\footnote{For readability, outlier scores greater than $3$ times the IQR were grouped together and shown separately in the right plot, which shows  the proportion of records with outlier scores beyond $3*$IQR  for each anomaly type.}
At the application level, the outlier scores assigned to normal instances spread further, and start to mix with the outlier scores  assigned to T2 anomaly instances, hence decreasing the model's separation ability. At the global level, this trend is aggravated, as we can see in Figure~\ref{fig:lm4-ae-ad2-global-separation}.

To understand why,  Figures~\ref{fig:lm4-ae-app2-bursty} and \ref{fig:lm4-ae-app2-bursty-crash} show the outlier scores of the T1 and T2 traces of Application 2. The outlier scores assigned to some normal points in the T1 trace are 
in fact higher than the T2 anomalies, due to two reasons: 
(a) {\em Different contexts}: T1 and T2 traces were generated under different input rates, with the  rate increase in T1 events around 2.5 times higher than in the T2 events.
(b) {\em Noisy training data}: The normal data in T1 is ``noisy''. In fact, the normal records in the T1 trace that obtained higher outlier scores than the T2 anomaly exactly match the high processing delay  due to Spark checkpointing activities. This indicates that the model has failed to capture these activities as normal behavior.

The above analyses show that our dataset carries a great deal of \textit{variability} across traces (e.g., different input rates, concurrency among programs), and a small amount of \textit{noise}. Such  variability and noise make our dataset challenging for the three DL-based AD methods tested in this study (\textbf{R2}).

\begin{table}[t]
    \centering
    \footnotesize
    \begin{tabular}{|c|ccc|cccccc|}
        	\hline
	\rowcolor{LightCyan}
    	 {\bf AD1} & F1 & Prec & Rcl & \multicolumn{6}{c|}{Rcl for Anomaly Types T1$\rightarrow$T6}\\ \hline \hline
   		\textbf{LSTM} & 0.77 & 0.67 & 0.96 & 1.00 & 1.00 & 1.00 & 1.00 & 1.00 & 0.67 \\\hline 
   		\textbf{AE} & 0.59 & 0.54 & 0.76 & 1.00 & 0.88 & 1.00 & 0.57 & 1.00 & 0.31 \\ \hline
   		\textbf{BiGAN} & 0.28 & 0.90 & 0.19 & 0.59 & 0.00 & 0.00 & 0.10 & 0.17 & 0.06 \\ \hline
    	\hline
	\rowcolor{MedCyan}
    	 {\bf AD2} & F1 & Prec & Rcl & \multicolumn{6}{c|}{Rcl for Anomaly Types T1$\rightarrow$T6}\\ \hline \hline
   		\textbf{LSTM} & 0.38 & 0.67 & 0.29 & 0.42 & 0.11 & 0.55 & 0.16 & 0.60 & 0.10 \\ \hline 
   		\textbf{AE} & 0.52 & 0.54 & 0.60 & 0.97 & 0.15 & 0.67 & 0.40 & 1.00 & 0.15 \\ \hline
   		\textbf{BiGAN} & 0.17 & 0.90 & 0.10 & 0.30 & 0.00 & 0.00 & 0.06 & 0.17 & 0.00 \\ \hline \hline
	\hline
	\rowcolor{cyan}
    	 {\bf AD3} & F1 & Prec & Rcl & \multicolumn{6}{c|}{Rcl for Anomaly Types T1$\rightarrow$T6}\\ \hline \hline
   		\textbf{LSTM} & 0.29 & 0.67 & 0.20 & 0.27 & 0.04 & 0.37 & 0.10 & 0.58 & 0.08 \\ \hline 
   		\textbf{AE} & 0.51 & 0.54 & 0.57 & 0.96 & 0.06 & 0.62 & 0.37 & 1.00 & 0.14 \\ \hline
   		\textbf{BiGAN} & 0.14 & 0.90 & 0.08 & 0.23 & 0.00 & 0.00 & 0.05 & 0.17 & 0.00 \\ 
   		\hline \hline
   		
   	\hline
	\rowcolor{DarkCyan}
    	 {\bf AD4} & F1 & Prec & Rcl & \multicolumn{6}{c|}{Rcl for Anomaly Types T1$\rightarrow$T6}\\ \hline \hline
   		\textbf{LSTM} & 0.13 & 0.67 & 0.08 & 0.00 & 0.00 & 0.07 & 0.00 & 0.58 & 0.06 \\ \hline 
   		\textbf{AE} & 0.49 & 0.52 & 0.56 & 0.94 & 0.06 & 0.58 & 0.37 & 1.00 & 0.14 \\ \hline
   		\textbf{BiGAN} & 0.14 & 0.86 & 0.08 & 0.23 & 0.00 & 0.00 & 0.05 & 0.17 & 0.00 \\ \hline
    \end{tabular}
    \caption{\small Median anomaly detection results (FS$_{\text{custom}}$, AD1:4)}
    \label{tab:custom-ls4-ad-detection}
    \vspace{-0.3in}
\end{table}

(3) \textit{Anomaly Type Comparison}:
For different anomaly types, Table~\ref{tab:custom-ls4-ad2-separation} shows that at the global level, the best separated types are T1, T3 and T5 for LSTM and AE, and T1, T4 and T5 for BiGAN.
The good performance for T1 (bursty input) and T5 (driver failure) across all methods are largely due to the fact these types have very visible impacts on many of the features output by FS$_\text{custom}$, e.g., features relating to the input rate, application delays and memory usage for bursty input, and virtually all features for driver failure.
However, most methods offer poor separation for T6 (executor failure) anomalies, due to the limited impact such anomalies have on the FS$_\text{custom}$ features, where the 6 executor features are  \emph{averaged across active executor spots}. As such, the impact of an executor going down is only  visible during the (short) period of time for which it shuts down and is potentially replaced. 
The above discussion shows that the variety of our anomaly types  present signals of different strength levels in the data. They offer challenges for designing, as well as  opportunities for analyzing, different AD methods, and feature engineering in the AD method will play 
a key role in preserving the signals for each anomaly type (\textbf{R3}).

(4) \textit{Method Comparison}: Regarding the separation ability, the best performing method is AE, followed by BiGAN, then LSTM for all levels. 
AE (and BiGAN) typically produce \emph{smooth} point-wise outlier scores, by taking averages over overlapping windows. The outlier scores produced by the LSTM, however, often exhibit discontinuous \emph{spikes} (see Appendix~\ref{appendix:ad-results}). For the task of range detection (AD2), such frequent mixes of high and low values make it hard to produce continuous ranges of high outlier scores, penalizing recall when the outlier threshold is set high or precision when the threshold is set low. 
Hence, we observe differences among AD methods as follows: for range detection (AD2), AE works the best while LSTM is the worse, mostly because the non-smooth outlier scores of LSTM make it hard to handle range anomalies (\textbf{R4}).

\minipa{Experiment 2 (FS$_{\text{custom}}$, AD2).} 
We next examine how the separation abilities translate into actual AD performance via threshold selection. Detection metrics for AD2 (range detection) are reported in the second section of Table~\ref{tab:custom-ls4-ad-detection}.
To generate the results, we ran each of the STD, IQR and MAD thresholding techniques, leading to different AD performance results for each AD method, for which we report the median performance in Table~\ref{tab:custom-ls4-ad-detection}.

Among the three methods, AE provided the best median F1-score, due to its best separation ability reported in the previous experiment. 
However, this F1-score of AE is not very high (0.52).
This is due to the difficulty in choosing a single threshold $T$ on the outlier score for all traces and anomaly types in an {\em unsupervised}  setting, where we select $T$ by using part of the training data, $D^2_{\text{train}}$. 
Figure~\ref{fig:lm4-ae-ad2-m-test-scores} shows the distribution of the outlier scores assigned  to the $D^2_{\text{train}}$ samples (the 3\% largest were cut for readability), along with the best threshold found on them. This threshold is then used  to flag anomalies in the test (disturbed) traces, as shown in Figure~\ref{fig:lm4-ae-ad2-global-separation}. We see that the anomalies whose scores lie left to $T$ will be missed, penalizing recall, and the normal records whose scores lie right to $T$ will lead to false positives, hurting precision. 
The above discussion shows that besides data characteristics, our benchmark poses another challenge on AD methods due to the requirement of unsupervised threshold selection (\textbf{R5}). 

\begin{figure}[t]
	\begin{center}
		\leavevmode
		\includegraphics[width=1.0\columnwidth,height=2.5cm]{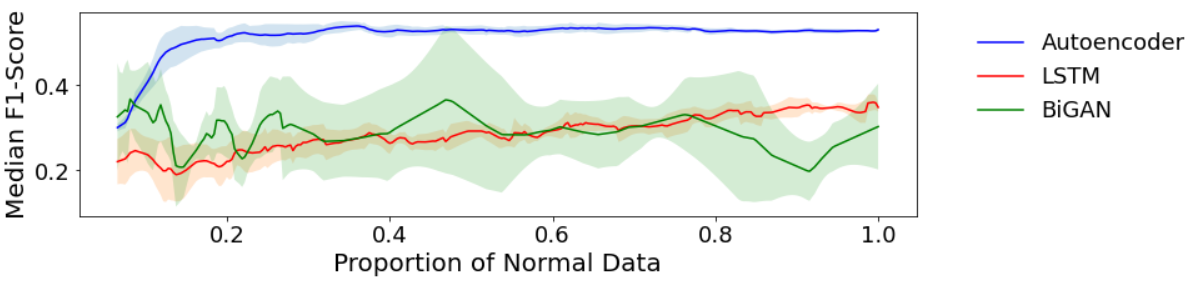}
	\end{center}
    \vspace{-0.1in}
	\caption{\small Median F1-score as training data increases for the three methods (FS$_{\text{custom}}$, AD2)} 
	\vspace{-0.25in}
    \label{fig:data-experiment}
\end{figure}

\minipa{Experiment 3 (FS$_{\text{custom}}$, AD2).}
To better understand the reasons behind the low  F1-scores, we study the effect of the amount of training data on each method in Figure~\ref{fig:data-experiment}. 
The amount of training data was varied by starting from the largest undisturbed trace, and then randomly adding one undisturbed trace at a time until reaching the full set of undisturbed traces (except for the BiGAN method for which multiple traces could be added at once due to its longer training time).
For each method, the above process was repeated 5 times. The average performance is reported in Figure~\ref{fig:data-experiment} using solid lines, while the shaded areas correspond to the confidence region with width of one standard deviation.

We can see that the three methods behave quite differently as training data increases. First, the AE method benefits a lot from the first few traces that it obtains for training, but quickly reaches a performance plateau afterwards. This seems to indicate that what is holding back the AE performance is not simply the lack of training data, but rather the actual challenges posed by our benchmark (see remarks R2, R3, R5).
On the other hand, the LSTM method and the BiGAN method to some extent seem to require more data to perform well. While LSTM exhibits roughly a linear trend, BiGAN appears less stable, which probably arises from the fact that GANs typically require more manual and calibrated tuning in order to converge to a good solution. For both these methods, adding more data could be beneficial, along with more tuning and experimentation to try to improve performance. 
Overall, this experiment suggests that the F1-score observed for AE could be primarily due its technical limitations for handling complex data, extracting most informative features for different anomaly types, and unsupervised threshold selection, while LSTM and BiGAN can further benefit  from more training data and extensive hyperparameter tuning (\textbf{R6}). 

\minipa{Experiment 4 (FS$_{\text{custom}}$).}
We next evaluate the AD methods under different AD levels, AD1-4, of our benchmark. 
Results are reported in Table~\ref{tab:custom-ls4-ad-detection}. 
(AD1)~Given our range-based anomalies, a good recall score is  easier to reach under AD1. We observe a general increase in performance for all methods. LSTM becomes the best method,  because its  spikes inside a real anomaly range are now sufficient for getting a good recall score, while using a high threshold to ensure good precision.
(AD3)~As AD3 awards less recall scores for late detection, AE  maintains its performance, indicating that its reported range anomaly is not concentrated at the end of the true range. For LSTM, the performance drops because the early detections it makes are more scattered and hence weigh less in recall. 
(AD4)~In AD4, reporting the same anomaly multiple times reduces the recall score. AE and BiGAN can maintain their performance while  LSTM degrades significantly. Again, the tendency of LSTM to produce  outlier scores in discontinuous spikes makes it more likely to report multiple anomalies where only one is needed.
Hence, we see that the different  AD levels in our benchmark indeed pose varying levels of challenges to  the AD methods (\textbf{R7}).

\ssubsection{ED Evaluation Results and Discussion}

\begingroup
\setlength{\tabcolsep}{2pt}
\begin{table*}
	\scriptsize
	\begin{center}
		\begin{tabular}{|c|ccccccccccc|ccccccccccc|ccccccc|}
			\hline
			&  \multicolumn{11}{c|}{\textbf{MacroBase}}  &  \multicolumn{11}{c|}{\textbf{EXstream}} 
			&  \multicolumn{7}{c|}{\textbf{LIME}} \\ \hline \hline
			& \multicolumn{2}{c}{Concise} & \multicolumn{2}{c}{Consistency} & \multicolumn{2}{c}{Norm.Cons} & \multicolumn{2}{c}{Prec} & \multicolumn{2}{c}{Rel} & Time (sec)
			& \multicolumn{2}{c}{Concise} & \multicolumn{2}{c}{Consistency} & \multicolumn{2}{c}{Norm.Cons} & \multicolumn{2}{c}{Prec} & \multicolumn{2}{c}{Rel} & Time (sec)
			& \multicolumn{2}{c}{Concise} & \multicolumn{2}{c}{Consistency} & \multicolumn{2}{c}{Norm.Cons} & Time (sec)	
			\\ 
			& ED1 & ED2 & ED1 & ED2 & ED1 & ED2 & ED1 & ED2 & ED1 & ED2 & ED1/2 & ED1 & ED2 & ED1 & ED2 & ED1 & ED2 & ED1 & ED2 & ED1 & ED2 & ED1/2 & ED1 & ED2 & ED1 & ED2 & ED1 & ED2 & ED1/2 \\
			T1 &  2.36&2.10&1.41&2.12&1.23&2.06&0.96&0.81&0.92&0.84&0.87& 2.25&2.17&1.59&3.36&1.56&4.74&0.88&0.82&0.75&0.45&0.0162&3.58&5.86&3.04&3.82&2.33&2.41&259\\ 
			T2 &  1.94&1.71&0.98&1.33&1.11&1.46&0.97&0.79&0.97&0.90&1.05& 3.89&3.71&2.13&3.59&1.37&3.25&0.71&0.87&0.53&0.51&0.0087&3.74&9.29&3.42&4.16&2.88&1.93&237\\ 
			T3 & 6.08&4.83&2.98&3.07&1.37&1.74&0.97&0.88&0.91&0.51&8.75& 3.35&3.75&2.13&3.67&1.57&3.40&0.82&0.78&0.65&0.18&0.0156&3.83&6.58&3.28&3.93&2.59&2.32&254 \\
			T4 & 1.51&1.55&1.11&2.90&1.59&4.82&0.73&0.52&0.77&0.42&0.14& 3.41&3.80&1.80&3.78&1.27&3.62&0.61&0.51&0.33&0.11&0.0106&4.03&6.10&3.34&3.97&2.56&2.57&241 \\
			T5 &  4.63&1.71&2.79&2.79&1.77&4.04&0.13&0.29&0.11&0.40&0.96& 1.66&1.71&1.08&2.52&1.47&3.35&0.34&0.40&0.34&0.41&0.0067&4.57&4.33&3.53&3.72&2.55&3.04&242\\
			T6 &  2.42&1.75&1.55&2.22&1.29&2.66&0.80&0.17&0.82&0.50&0.24& 2.75&2.88&1.20&3.50&1.30&3.94&0.72&0.30&0.59&0.27&0.0088&4.20&4.43&3.44&3.65&2.66&2.84&239\\ \hline \hline
			\rowcolor{MedCyan}
			Ave &  3.16&2.28&1.80&2.40&1.39&2.80&0.76&0.58&0.75&0.59&2.00& 2.88&3.00&1.66&3.41&1.42&3.72&0.68&0.61&0.53&0.32&0.0111&3.99&6.10&3.34&3.88&2.60&2.52&245\\ 
			\hline
		\end{tabular}
		\caption{\small Results of ED methods, MacroBase, EXstream, and LIME, in terms of conciseness, consistency, accuracy, and running time}
		\vspace{-0.3in}
		\label{table:ed_result}
	\end{center}
\end{table*}
\endgroup

\begin{figure}[t]
	\centering
	\begin{tabular}{lcc}
		\subfigure[\small{EXstream}]
		{\label{fig:exstream_example}\includegraphics[height=2.2cm]{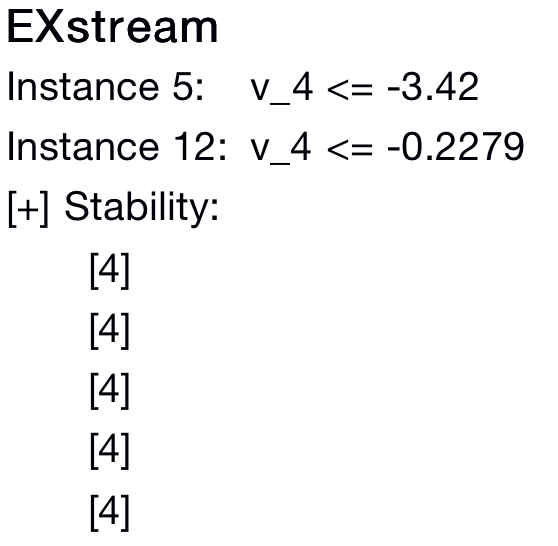}}
		&
		\subfigure[\small{MacroBase}]
		{\label{fig:macrobase_example}\includegraphics[height=2.2cm]{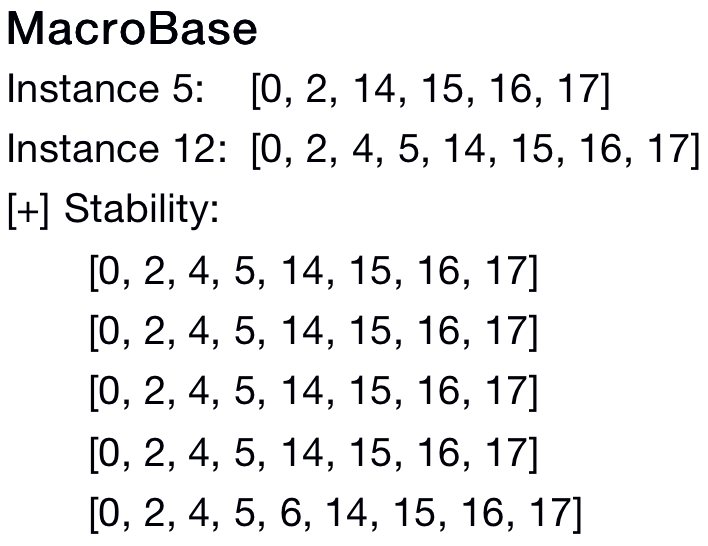}}
		&
		\subfigure[\small{LIME}]
		{\label{fig:lime_example}\includegraphics[height=2.2cm]{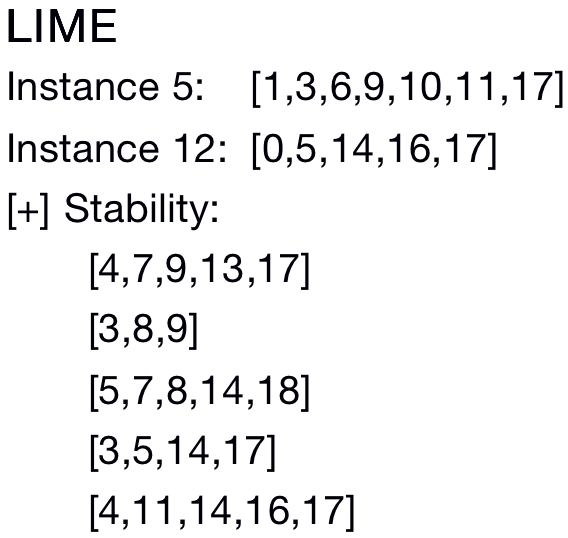}}
	\end{tabular}
	\vspace{-0.2in}
	\caption{\small Example explanations given by EXstream, MacroBase, and LIME for two instances of stalled input (T3) anomaly}
	\label{fig:ed_example_all}
	\vspace{-0.2in}
\end{figure}

Next, we report results of running MacroBase~\cite{macrobase:sigmod17}, EXstream \linebreak ~\cite{ZDM2017}, LIME~\cite{Ribeiro0G16} on our benchmark dataset to generate anomaly explanations. For each anomaly, MacroBase and EXstream tried to explain its separation from a reference dataset, while LIME tried to explain the reason behind the high record outlier scores assigned by an AD model (AE in our case). Since LIME only explained the predictions of window size $s$ of our AE model, if an anomaly was larger than $s$, we created multiple windows for LIME to explain.

Table~\ref{table:ed_result} summarizes {\em conciseness}, {\em consistency}, {\em normalized consistency}, {\em accuracy}, and {\em running time} for local (ED1) or global (ED2) explanations using the three ED methods.
We also show example explanations in Figure~\ref{fig:ed_example_all}, which are the explanations returned for two instances of stalled input anomaly (T3). Only feature indices are reported here (see Appendix~\ref{appendix:custom-set} for the feature names). Figure~\ref{fig:exstream_example} reports the complete explanations returned by EXstream, while showing the features appearing in the explanations for the others due to limited space. For each method, we also show the features returned when explaining 5 different samples of anomaly instance \#12 (stability). 
	
\noindent
\textbf{Explanations.}
The explanations shown highlight the impact of feature correlation. Although MacroBase and EXstream output different features (Figure~\ref{fig:exstream_example} and ~\ref{fig:macrobase_example}), they might both be correct. For example, features 4, 5 and 14 are related to processed records, received records, and CPU time (see Appendix~\ref{appendix:custom-set}). For a human user, it makes sense for these three features to be used for a stalled input anomaly. EXstream picks up only one most important feature among the correlated ones. MacroBase returns all important features no matter whether they are correlated or not.
LIME is known to have inconsistency issues~\cite{Molnar-book}, which is also illustrated here.

\noindent
\textbf{Algorithm Analyses.} We discuss the results for each algorithm.

\noindent
\underline{MacroBase.} 
(1)~MacroBase generated explanations of 3.16 features on avg. across anomaly types. For some anomaly types, e.g., T3, it generated longer explanations (6-7 features). 
The algorithm does not consider compactness, outputting longer explanations in presence of correlated features.
(2)~The explanations it provided were not very locally stable (ED1 consistency), with an entropy score outside the ideal range of $H_1=0$ and $H_3=1.58$. This relates to a correlation between conciseness and stability: as Table~\ref{table:ed_result} shows for different anomaly types, longer explanations tend to be less stable. For global consistency, its concordance value further degrades, using inconsistent features for explanations of the same anomaly type. This issue was alleviated by measuring normalized consistency. In the example of Fig.~\ref{fig:macrobase_example}, the stability of MacroBase for instance \#12 is 3.09, while its normalized stability is 1.03, almost perfect. The latter value is in accordance with human intuition, since across the 5 runs, the reported features were almost the same.
(3)~Its ED1 accuracy is good for some anomaly types (e.g., T1-3), but poor for some others (e.g., T5). As expected, its accuracy degraded from ED1 to ED2. 
(4)~Its execution time is around 1 sec, except for T3.

\noindent
\underline{EXstream.}  
(1)~EXstream provided more concise explanations, using multiple techniques to prune marginally related features. 
(2)~It achieved good explanation stability, with an average entropy score of $H_3=1.58$, but worse explanation concordance, be it normalized or not. This is likely due to our exclusion of the false positive filtering step, requiring additional labels. Thus, some features standing out as different during anomalous periods might not be related to the anomaly but to normal changes between two contiguous periods. Such features are likely to be more distinct between instances of different contexts than for perturbations of the same instance, hence the greater effect on concordance.
(3)~Its ED1 accuracy is good for some anomaly types (e.g., T1-3), but not for others (e.g., T3-6), especially in recall. Its accuracy degraded for ED2, as expected.
(4)~Its execution time is very short, being a streaming algorithm. 

\noindent
\underline{LIME.} 
(1) LIME generated longer explanations, e.g., for anomaly types T1-T3. 
(2) Its consistency scores were better maintained from ED1 to ED2, with normalized metrics alleviating the observed correlations between size and consistency.
Accuracy measures do not apply to LIME, since it could not be compared to the others for prediction.
(4)~Its execution time is very long, up to 245 sec on avg.

\noindent
{\bf Comparison.}
We next compare the three methods, first in {\em conciseness} and {\em stability}. MacroBase lacks a mechanism for minimizing the size of an explanation, while LIME relies on sparse linear regression to select few features. Neither was as effective as EXstream, which eagerly prunes marginally related features through its heuristic to a non-monotone submodular optimization problem. Stability being positively correlated to conciseness, the longer explanations of MacroBase and LIME are also less stable. 
For global explanations, {\em concordance} was harder to achieve than stability for all methods, i.e.,  a user is likely to see explanations built on different features for anomalies of the same type, which is undesirable.
This lack of concordance indicates a direction for future ED research.
For local {\em accuracy}, the logical formulas derived by MacroBase and EXstream on a subset of each instance could be evaluated for AD on their remaining part and neighboring normal data. MacroBase was more accurate, paying the cost of evaluating a large number of feature combinations, while EXstream suffered in recall, ``overfitting'' each anomalous instance. 
LIME, returning only feature importance scores, could not be evaluated for accuracy.
In the global setting, accuracy degraded for all methods. By hard-coding context-dependent constants in predicates, the explanations of MacroBase and EXstream did not generalize well to other contexts (e.g., different input rates). This phenomenon is intrinsic to point-based explanations.
In order to free explanations from such context-dependent predicates, transitioning from point-based to temporal explanations, capturing causal and context-free relationships between events, could be a direction for future research. 
In {\em efficiency}, EXstream was the fastest, taking \~0.01 sec to generate explanations, against 0.2-9 sec for MacroBase and $>$ 4 min for LIME. As such, LIME is unsuitable for stream processing use cases, with its high latency preventing timely corrective actions, e.g., avoiding an application crash or denial of service. 

\vspace{-0.1cm}
\section{Conclusions and Future Directions} \label{sec:conclusions}

In this paper, we presented \system\ -- a novel public benchmark for explainable AD, and demonstrated its utility through an experimental analysis of selected AD and ED algorithms from recent literature. Our AD results show that \system's dataset is valuable for evaluating AD algorithms due to rich signals and diverse anomaly types included in the data. Yet more importantly, our results reveal the limitations of these AD methods for semi-supervised learning under noisy training data and 
mixed anomaly types. 
On the ED front, the literature lacked comparative analysis tools and studies. Our benchmark fills this gap by providing a common framework for analyzing the strengths and limitations of diverse ED methods in their conciseness, consistency, accuracy, and efficiency.
These results call for new research to advance the current state of the art of AD and ED, as well as integrated solutions to anomaly and explanation discovery. For a true integration, ED methods should first become capable of discovering range-based explanations, which is also a key step towards automated root cause analysis (a.k.a., ``why explanations''). \system's dataset and extensible design are well-positioned to support research progress towards these goals in the long term. Going forward, we envision \system\ to develop into a collaborative community platform for fostering reproducible research and experimentation in the area. We intend to actively maintain and extend this platform, as well as welcoming feedback and contributions from the AD and ED communities.

\balance
\bibliographystyle{ACM-Reference-Format}
\bibliography{refs,ed-refs}

\appendix
\section{Public Repository}

The dataset, code, and documentation for \system\ are publicly available at \url{https://github.com/exathlonbenchmark/exathlon}. 

\section{Details on Data Collection}
\label{appendix:data}

In this section, we provide additional details on our anomaly design and data collection. 

\begin{figure*}[t]
	\centering
	\begin{tabular}{lcc}
		\hspace{-0.1in}
		\subfigure[\small{Metrics in an undisturbed trace}]
		{\label{fig:normal}\includegraphics[width=0.33\textwidth,height=3.6cm]{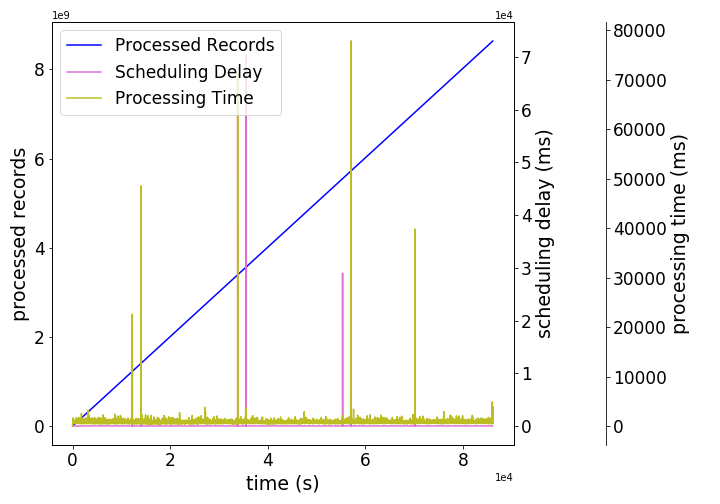}}
		&		
		\hspace{-0.1in}
		\subfigure[\small{Bursty input}]
		{\label{fig:burstyinput2}\includegraphics[width=0.33\textwidth,height=3.5cm]{figures/bursty_input.png}}
		&
		\hspace{-0.1in}
		\subfigure[\small{Bursty input until crash}]
		{\label{fig:burstycrash}\includegraphics[width=0.33\textwidth,height=3.5cm]{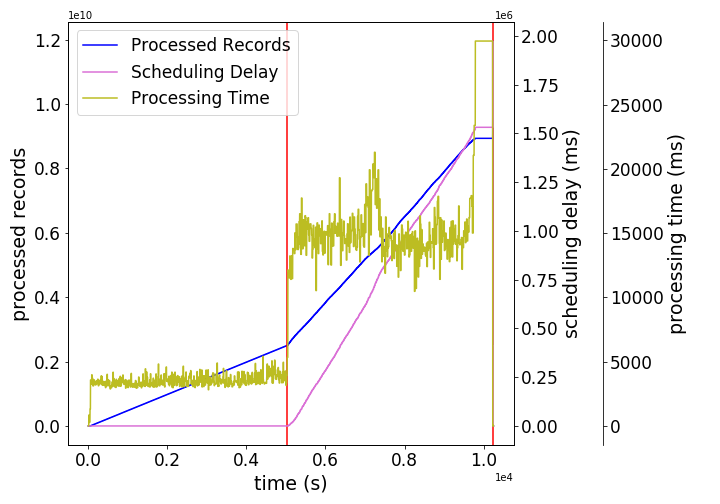}}
		\\
		\hspace{-0.2in}
		\subfigure[\small{Stalled input}]
		{\label{fig:stalledinput}\includegraphics[width=0.33\textwidth,height=3.5cm]{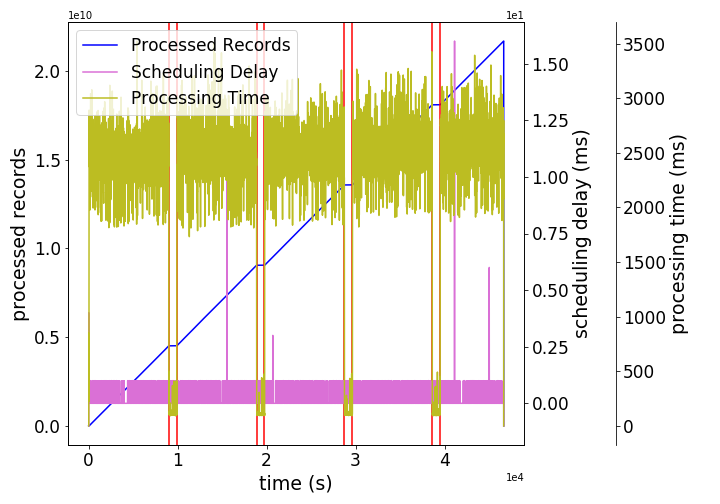}}
		&
		\hspace{-0.1in}
		\subfigure[\small{CPU contention}]
		{\label{fig:cpucontention}\includegraphics[width=0.33\textwidth,height=3.5cm]{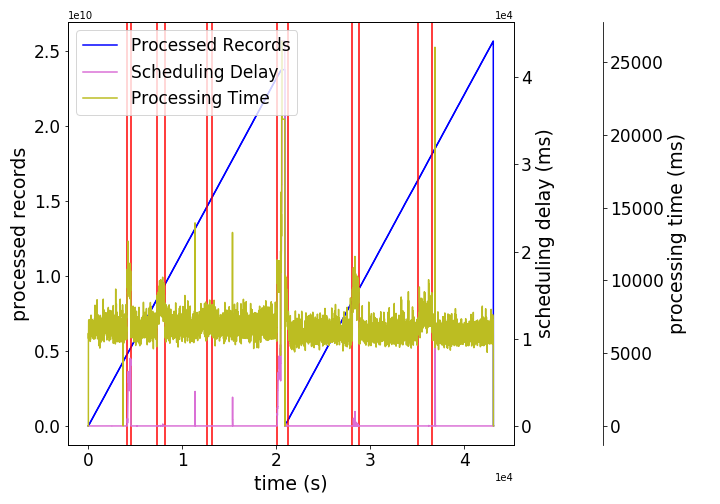}}
	\end{tabular}
\vspace{-0.2in}
\caption{\small Metrics observed in normal data and anomaly instances (a pair of red vertical bars marks a root cause event) }
\vspace{-0.1in}
\end{figure*}

\subsection{Spark Workload}
\label{appendix:spark-applications}

Our Spark workload consists of 10 stream processing applications that analyzed user click streams from the WorldCup 1998 website~\cite{LiMD+12} (replicated with a scale factor for our long-running applications). The applications processed data formatted as \texttt{(user\_id - - timestamp - - url)} records, using a batch interval of 5 seconds. Join data stored in HDFS was also available for the applications to use, in the form of \texttt{(page\_rank - - url)} records.

We summarize the operation performed by each application below, and provide their source code under \url{https://github.com/exathlonbenchmark/exathlon/tree/master/apps} for more details: 

\noindent \textbf{App 1.} GROUP BY + COUNT last batch values.

\noindent \textbf{App 2.} GROUP BY + COUNT + FILTER values from the start.

\noindent \textbf{App 3.} GROUP BY + COUNT + FILTER values in a 30s sliding window (with 20s steps).

\noindent \textbf{App 4.} GROUP BY values in a 30s sliding window.

\noindent \textbf{App 5.} JOIN + GROUP-BY + SUM values in a 30s sliding window.

\noindent \textbf{App 6.} JOIN + GROUP-BY + SUM values in a 30s sliding window.

\noindent \textbf{App 7.} GROUP-BY + SIMULATED UDF for values in a 30s sliding window.

\noindent \textbf{App 8.} GROUP-BY + COUNT + FILTER values in a 10s jumping window.

\noindent \textbf{App 9.} GROUP-BY values in a 10s jumping window.

\noindent \textbf{App 10.} GROUP-BY + COUNT + FILTER values in a 10s jumping window.

The results for these operations were entirely saved on HDFS for applications 6, 8, 9 and 10. For other applications, the results were saved for the first batch only.

Overall, these applications commonly involved group-by aggregations (counts or sums), and varied in terms of %
window type (sliding vs. jumping), parameters (size and slide) and filtering conditions. 
Two of the applications additionally involved join operations, and one of them used a user-defined function. As such, our workload aimed to capture a diverse mix of popular streaming primitives.

\subsection{Undisturbed and Disturbed Traces}
\label{appendix:traces}

\noindent
{\bf Undisturbed Traces.}
To collect traces characterizing the normal execution behavior of our Spark cluster, we continually ran experiments (5/10 randomly selected applications at a time) for a month, using input rate values and Spark parameters that suited the capacity of our cluster. Some traces affected by cluster downtime were manually pruned, leaving 59 undisturbed traces, around 15.3GB of data, to constitute the main bulk of our dataset. We consider these traces to be representative of the normal state because the cluster was not interrupted or stress tested by disruptive external events during their recording.

It is important to note that even undisturbed traces exhibit a great deal of variability in the recorded metrics.
For example, Figure~\ref{fig:normal} shows that both the {\em scheduling delay} (delay between the scheduling of a task and the start of its processing) and {\em processing time} (time taken to process an input batch of streaming data) of the recorded application could rise high beyond their usual ranges of values, while its total {\em number of processed records} steadily increased over time. 
The spikes observed for the first two metrics were likely caused by other system operations, e.g., checkpointing in Spark or CPU usage by a DataNode in HDFS (Hadoop File System). Since such variations appear in almost every trace, they should be considered to be part of the normal state. 

\noindent
{\bf Disturbed Traces.} 
Disturbed traces were obtained by introducing anomalous events during an execution. Based on discussions with industry contacts from the Spark ecosystem, we came up with 6 types of anomalous events. When designing these, we considered that:
(i) they lead to a visible effect in the trace,
(ii) they do not lead to an instant crash of the application (since AD would be of little help in this case),
(iii) they can be tracked back to their root causes.

\minipb{Bursty Input (Type 1):} 
For every application, the user expects a certain range of data input rates and configures Spark parameters to allocate sufficient resources for such input rates.  However,  on some special occasions (e.g., a special sales day for a retailer) the input rate can  increase significantly, and existing resources be insufficient for handling the higher data rate. Such phenomena can be reflected in the data, e.g., in increased values of processing time, memory usage, and scheduling delay, as shown in Figure~\ref{fig:burstyinput2}. It is because each batch of data (e.g., received in the past 10 seconds) increases dramatically in size, and the processing time of the batch increases accordingly. When the processing time exceeds the batch interval (e.g., 10 seconds), the application cannot process data fast enough; so data is put in memory by the receivers and such data experiences a higher delay before it is scheduled for processing. Early detection of bursty input is helpful because it can lead to corrective actions such as allocating more resources to the application.

To mimic input rate spikes, we ran a disruptive event generator (DEG) on the Data Senders to temporarily increase the input rate by a given factor for a duration of 15-30 minutes and then set the input rate back to its normal range. We repeated this pattern multiple times during a given trace, with sufficient gap between two instances. As such, we created a total of 29 instances of this anomaly type over 6 different traces.

\minipb{Bursty Input Until Crash (Type 2):} 
Spark developers rank the out of memory (OOM) condition to be the number one reason for Spark application failures. To capture such phenomena, we extended the bursty input design:  instead of reducing the input rate back to a normal range, we kept the input at the same high rate until the application eventually crashed. Figure~\ref{fig:burstycrash} shows how the system behaviors during such an event are reflected in the data: the application starts with a normal input rate. At some point, the input rate rises high, and the processing time and scheduling delay build up until one of the executors fails due to an out-of-memory error. 
A new executor is then launched to replace the failed one, but the sustained high rate causes more executors to fail. When the number of failed executors reaches a threshold, the application is killed by Spark, constituting an application crash. 

To instrument this anomaly type, we ran the DEG  forever, first crashing the executors due to lack of memory, and eventually crashing the application. We injected this anomaly into 7 different traces. 

\minipb{Stalled Input (Type 3):} 
Another type of input-related anomaly is stalled input, i.e., no data is injected into the Spark application. This may  indicate a failure of the data source (e.g., Kafka or HDFS) and results in a waste of resources as the driver and executors are still running with no data to process. As shown in Figure~\ref{fig:stalledinput}, this type of anomaly can  be reflected in our collected metrics: since no data is coming from the receiver, the number of processed records, as well as other related metrics, do not increase, and the processing time is much lower than in the normal state.

To generate these anomalies, we ran a DEG that set the input rates to 0 for about 15 minutes, and then periodically repeated this pattern every few hours. This gave us a total of 16 anomaly instances across 4 different traces..

\minipb{CPU Contention (Type 4):} 
A Spark cluster is usually run with a resource manager (e.g., YARN) that allocates resources to each application so that each of them has exclusive access to its own resources, including CPU and memory. However, the resource manager cannot prevent external programs from using the resources that it allocated previously, which is a common reason for performance issues in distributed environments with high concurrency. For instance, it is not uncommon to have a Hadoop DataNode using a large amount of CPU on a node also used by a Spark application. This generates a contention for resources that can often slow down the progress of the Spark application.

We reproduced this type of anomaly using a DEG that ran Python programs to consume all CPU cores available on a given Spark node. We created 26 such anomaly instances over 6 different traces. An example trace containing multiple instances of this type of anomaly is shown in Figure~\ref{fig:cpucontention}.
From this, we see there are several intervals during which the processing time is higher than normal due to  CPU contention on the affected node, meaning that one or more executors are using fewer resources than allocated. If the processing time is still lower than the batch interval, the CPU contention does not affect much the progress of the application;  otherwise, scheduling delay starts to build up. Besides processing time, CPU contention can have a more severe impact on the Spark application: if the driver is performing a lot of computation and CPU contention occurs on the driver node, it can cause it to crash.
Then the application master has to restart the driver, which will itself restart all the executors.

\minipb{Driver Failure (Type 5) and Executor Failure (Type 6):}
Other types of anomaly include abrupt failures in distributed systems. A common example is a node failure caused  by a  hardware fault or a maintenance operation. In this case, all the processes (drivers and/or executors) located on that node become unreachable. Such processes must be restarted on another node, which causes processing delays. 
We created such anomalies by failing driver processes, causing cumulative metrics to drop back to their initial values and the application to stop until its driver is restarted in about 20 seconds.
We also created anomalies by failing executor processes, which get restarted 10 seconds after the failure, but whose effects on metrics such as processing delay may continue longer. 
We created 9 driver failures and 10 executor failures over 11 different traces.

\subsection{Extended Effect Intervals}
\label{appendix:extended-effect}

In this section, we describe the way that we set the extended effect interval (see Table~\ref{tab:traces}) for each anomalous event.
Each anomaly instance was initially labeled with its known type and {\em root cause interval} (RCI). 
However, we observed that some injected events could have a long-lasting effect on relevant metrics such as processing time and scheduling delay even after their end time.
To characterize such events, we therefore sought to extend their root cause interval to include this long-lasting effect.
More specifically, we used domain knowledge to set an {\em extended effect interval} (EEI) for each anomalous event, starting immediately after the end of its RCI, and ending
at a point after which we deem anomaly detection not helpful.
In practice, we either set the end of an EEI to when the application had fully restarted, or to when its main metrics had come back to normal. Here are the rules we used for each type of anomalous event: 

\begin{enumerate}
\item Bursty Input: The end of the EEI is set to the point when highly related metrics such as the processing time and scheduling delay come back to normal.

\item Bursty Input Until Crash: The EEI is set to null, because the root cause event already ends at the time of the application crash. 

\item Stalled Input: The end of the EEI is set to the point when the processing time comes back to normal (the application restarts processing data at its usual rate).

\item CPU Contention: 
a) If the effect is increased processing time, we set the end of the EEI to the time when processing time and scheduling delay come back to normal.
b) If the effect is a driver crash, we set the end of the EEI to the time when the application restarts (typically 1 minute after the crash in practice). 

\item Driver Failure:
The end of the EEI is set to when the application restarts.

\item Executor Failure:
If the executor is replaced, we observe an increase in scheduling delay during the replacement time. If it is not replaced, then the buildup in scheduling delay either lasts temporarily or until the application crashes. In all these cases, we set the end of the EEI to the point when the scheduling delay comes back to normal.
\end{enumerate}

\noindent The combined anomaly interval, RCI plus EEI, gives our benchmark more freedom to evaluate anomaly detection algorithms.  

\section{Details on Benchmark Design}
\label{appendix:benchmark}

\begin{table}[t]
\centering
{\small
\begin{tabular}{|l|c|c|c|c|c|c|}
\hline
& \multicolumn{3}{c|}{\bf Precision} & \multicolumn{3}{c|}{\bf Recall} \\
& \multicolumn{3}{c|}{\bf Parameters} & \multicolumn{3}{c|}{\bf Parameters} \\
\hline
{\bf AD Functionality Level} & {\bf $\alpha$} & {\bf $\delta$} & {\bf $\gamma$} & {\bf $\alpha$} & {\bf $\delta$} & {\bf $\gamma$} \\
\hline
\hline
AD1: Anomaly Existence & 0 & Flat & 1 & 1 & N/A & N/A \\
\hline
AD2: Range Detection & 0 & Flat & 1 & 0 & Flat & 1 \\
\hline
AD3: Early Detection & 0 & Flat & 1 & 0 & Front & 1 \\
\hline
AD4: Exactly-Once Detection & 0 & Flat & 0 & 0 & Front & 0 \\
\hline
\end{tabular}
}
\caption{\small Range-based precision/recall parameter settings}
\label{tab:params}
\end{table}

\subsection{Anomaly Detection (AD) Functionality}
\label{appendix:ad-functionality}

\noindent
\textbf{Range-based Precision and Recall.}   
Table \ref{tab:params} further shows how to set the tunable parameters of the range-based precision/recall framework~\cite{TatbulLZAG18} to capture the requirements of AD1-AD4 in general. 
We provide a brief insight for these parameters here, and refer the reader to the original paper for further details \cite{TatbulLZAG18}. 
The $\alpha$ parameter is to reward detecting the existence of an anomaly range and is only relevant here in the context of recall when we only care about anomaly existence. Therefore, it always gets a 0 value except when computing AD1's recall. 
The $\delta$ parameter captures the positional bias, i.e., the reward to be earned from which portion of the anomaly range was successfully detected. It is the most relevant for AD3 and beyond, where detection latency is considered. For recall of AD3-4, $\delta$ should be set to ``Front'' to favor early detection, and ``Flat'' in all other cases, where detection position does not matter. 
Finally, the $\gamma$ parameter is to penalize fragmented ranges (i.e., an anomaly range detected as multiple subranges) and is, therefore, directly relevant to exactly-once detection. For AD4, we set $\gamma$ to 0, indicating that fragmentation is strictly undesirable; for all others, it has no penalizing effect to the score. 
We would like to note that there is a fourth parameter, $\omega$, which is to compute how much reward is earned from the size of the anomaly range that is correctly detected (relevant for AD2 and beyond). We use its default, additive definition in the original model with a minor normalization adjustment to ensure monotonicity. More specifically, we make the precision/recall scores decrease from AD2 to AD3 by defining the union of detected ranges under AD2 as the best positioned (i.e., the earliest) one under AD3 to achieve a monotonic behavior.

\noindent
\textbf{Learning Settings.}   
By default, our benchmark offers only undisturbed traces  as training data for building an AD model, because it is the most practical for real-world use. 
This can be seen as a {\em ``noisy'' semi-supervised anomaly detection} problem~\cite{ChandolaBK2009}, meaning that the training data is assumed mostly normal but subject to small amounts of noise.
In this context, \system\ also tests how well a learned AD model generalizes to different workload characteristics, including the Spark application ($A$), input rate ($R$), and concurrency ($C$) characteristics. 
We offer four alternative learning settings, considering the following two aspects:

\noindent
\underline{\emph{Modeling Subject (1-App vs. N-App Learning)}:} Spark applications differ in workload characteristics (e.g., CPU intensive versus I/O intensive). Such characteristics may lead to different runtime observations (e.g., a CPU-intensive application would be more sensitive to CPU contention anomalies). Hence detecting anomalies in traces of different applications requires more generalization power.
(i)~\emph{1-App learning} focuses on training and evaluating an AD model on a single application basis. As such, there is no need for the model to generalize to other applications.
(ii)~\emph{N-App learning} trains an AD model using multiple applications together, and hence requires it to jointly learn how to characterize different applications.

\noindent
\underline{\emph{Training Constraints (Many vs. Few Examples)}:}
Each of our traces contains data characterizing an application running with a given input rate in a given concurrent environment.
An AD model trained on undisturbed traces with certain ($A$, $R$, $C$) settings may later be subject to a new trace with a previously unseen ($A$, $R$, $C$) setting.
A well-learned model is supposed to generalize across these differing workload settings. 
However, such generalization power is easier to achieve if the training data includes many examples of ($A$, $R$, $C$), 
or many ($R$, $C$) examples in the {\em 1-App} learning setting. 
By default, \system\ reserves the disturbed traces entirely for testing.
As such, the training data includes only the 59 undisturbed traces, which is unlikely to cover most ($A$, $R$, $C$) values and hence constitutes the {\em Few Example}-training scenario. This  bears similarity with the few-shot learning problem~\cite{WangYKN20} recently studied in the ML community, and poses a great challenge for learning.
\system\ also offers a simpler, yet less realistic, training setting, {\em Many Example}-training, in case some algorithms cannot learn from limited examples. Here, we allow the AD algorithm to also leverage an earlier segment (with normal data only) of each disturbed trace in training, while testing on the anomalies in the later segments of the same traces. This way, the model is given a chance to ``peek'' at the normal state (including all workload characteristics) of a particular test trace.
This simulates the scenario of a model trained on many ($A$, $R$, $C$) examples, some of which similar to the current test trace.

By combining the above options, we obtain four learning settings to experiment with: {\em LS1: 1-App, Many-Examples}; {\em LS2: N-App, Many-Examples}; {\em LS3: 1-App, Few-Examples}; and {\em LS4: N-App, Few-Examples}. 

\subsection{Broader Applicability of AD Evaluation}
\label{appendix:ad-applicability}

\textbf{Applicability to Other AD Techniques.}
In its current form, the Exathlon benchmark considers all techniques that work by 1) training a model for data normality via learning a specific task on undisturbed traces, 2) assigning real-valued outlier scores to new test records as proportional to the error made by the model, and 3) deriving binary predictions for these records based on whether their scores exceeded a given threshold. In our experimental study, this third step was performed in the same way for the three compared methods (LSTM, AE, and BiGAN).

In practice, any technique that can assign an outlier score to each record of a test sequence based on a model learned from (mostly) normal data can directly be evaluated using this benchmark. Depending on the feature extraction step, the range of directly applicable techniques is hence quite broad, ranging from deep learning-based techniques like the ones presented in this paper to simpler statistical techniques~\cite{bianco01}, to other recent techniques for subsequence AD~\cite{seq2graph}, \cite{norm} and data drift detection~\cite{conf-constraints}. 

Although not considered in this paper, our benchmark could also be used for problems focusing on sliding windows in disturbed traces only, with or without pre-training a model on undisturbed traces. This would extend the range of applicable techniques to also include the ones of the distance-based outlier detection for data streams (DODDS) literature~\cite{Tran2015}. In such streaming scenarios, where both CPU time and memory usage can be of primary concern, the maximum memory used to update the outlier information could be reported along with the inference time as part of our AD efficiency metrics, since this metric is frequently encountered in the DODDS literature.

\textbf{Applicability to Other Datasets.}
The AD metrics of Exathlon could in principle be used with any labeled time series anomaly test datasets similar to ours. For datasets where ground truth anomalies are provided as ranges, like those most frequently found in the discord discovery literature (e.g., the simulated engine disks (SED) dataset and the MIT-BIH Supraventricular Arrhythmia Database (MBA) recently used in \cite{seq2graph}), the four AD levels of the benchmark would be directly usable. For datasets where labels are provided as points, our benchmark would still be usable by resorting to the classical definitions of precision and recall metrics, which are special cases of the range-based ones \cite{TatbulLZAG18}. The availability of labels for multiple anomaly types (e.g., as in MBA \cite{physiobank}, \cite{mba-impact}) would also enable to directly report type-wise recall scores in addition to the global ones (where all anomaly types are considered the same).

Data granularity is use case dependent. That is, trace/application-level data separation may not make sense for all use cases. Depending on the use case, there may be other granularity levels or only a single one. Our AD evaluation metrics and methodology are equally applicable to such use case scenarios.

While we have not applied our AD metrics to other anomaly datasets beyond our Spark-based traces, the range-based precision/recall framework constituting their foundation was effectively tested on multiple real and artificial time series anomaly datasets from the AD literature \cite{TatbulLZAG18}. In fact, this framework is a generalization of the classical precision/recall metrics from point-based data to range-based data. As such, it is a superset that encompasses the classical precision/recall metrics and other similar metrics based on them (e.g., F-score), which are widely used in the literature for AD evaluation. As was shown in \cite{TatbulLZAG18} based on real datasets from the Numenta Anomaly Benchmark \cite{numenta:icmla15}, the range-based metrics can in fact be tuned to simulate the behavior of both the point-based metrics and Numenta's AD scoring metrics, as well as capturing additional behaviors that different AD applications may require. The range-based metrics also provide a modular and extensible formulaic structure. As such, we believe that our choice of AD metrics both provide good coverage of the existing metrics in the literature as well as being amenable to future extensions if necessary.

\subsection{Broader Applicability of ED Evaluation}
\label{appendix:ed-applicability}

\textbf{Applicability to Other ED Works.} 
While a user study may be the best way to evaluate the usefulness of explanations, it is not always available and may come at a high cost. Therefore, our benchmark aims to provide {\em automated} evaluation of ED methods based on intuitive metrics. For selecting those, we surveyed multiple other ED works and summarized the main metrics they used (see Table \ref{tab:survey_measurement}). Among these, we found conciseness and accuracy to be the most general and intuitive: human users typically prefer small sets of important factors as explanations, and if these were to be used as an AD tool, they should ideally achieve good accuracy. We also added the concept of consistency, given that accuracy is not always applicable or meaningful for a given ED method. As an example, running LIME to explain the outlier score assigned to a given test sample might not be directly usable for binary prediction. Another issue might be the impact of hard-coded constants in explanation predicates. For example, a valid explanation to a human user could be that ``the CPU usage is too high''. However, this notion of ``too high'' might be a relative concept, meaning above 50\% in a given context but above 80\% in another. A lot of ED methods typically include such constants in their explanations, sometimes leading a ``good'' instance explanation to have poor accuracy when applied to other instances. Like conciseness, the concept of consistency has the benefit to rule out the effect of such constants in the evaluation of an ED method. Although the concept of these three metrics is common to all evaluated methods, the way we calculate them typically depends on how the method works. For instance, if the method takes a dataset as input to generate the explanation, like in~\cite{ZDM2017, macrobase:sigmod17, LakkarajuBL16, WuHPZ2018}, we consider the consistency implementation described in Section~\ref{sec:benchmark}. For other methods, e.g., ~\cite{Ribeiro0G16, Ribeiro0G18, LundbergL17}, the calculation of consistency is window-based, implemented for LIME~\cite{Ribeiro0G16} in our pipeline. If needed, other implementations can also be added by the user. 

\textbf{Applicability to Other Datasets.} 
The ED metrics of \system\ are directly applicable to the datasets used in~\cite{ZDM2017, macrobase:sigmod17}. Our metrics can also be applied for evaluating explanations of classification results, which is relevant for the datasets of~\cite{LakkarajuBL16, WuHPZ2018, Ribeiro0G18}, balanced and used for classification problems. For the text and image datasets used in ~\cite{Ribeiro0G16, LundbergL17}, our metrics are applicable after an additional data preprocessing step. If an application scenario does not provide any detailed anomaly type information, one can consider all anomalies as of the same type. Finally, if there is no clear separation of a dataset into instances, the user may need to decide to consider either ED1 or ED2 based on the dataset semantics.

\begingroup
\setlength{\tabcolsep}{1pt}
\begin{table}[t]
	\centering
	\footnotesize
	\begin{tabular}{|c|c|c|c|}
		\hline
		 ED Work & Metrics & Applicability of \system\ & Comment \\ \hline \hline
		\multirow{3}{*}{{EXstream~\cite{ZDM2017}}} & 
		{Conciseness} & {\textbf{Yes}} & {ED2 Conciseness} \\
		&  {Consistency} & {No} & {Needs ground truth}  \\
		& {Accuracy} & {\textbf{Yes}} & {ED2 Accuracy} \\ \hline
		{{MacroBase~\cite{macrobase:sigmod17}}} & 
		{Efficiency} & {\textbf{Yes}} & {Running time}  \\ \hline
		\multirow{2}{*}{{LIME~\cite{Ribeiro0G16}}} & 
		{Coverage of gold set} & {No} & {Model-dependent}  \\
		& {Overlap with noise} & {No} & {Model-dependent}  \\ \hline
		\multirow{2}{*}{{Anchors~\cite{Ribeiro0G18}}} & 
		{Coverage} & {No}& {Algorithm-specific}  \\
		& {Precision} & {\textbf{Yes}} & {ED1 Accuracy}  \\  \hline
		\multirow{3}{*}{{Decision S.~\cite{LakkarajuBL16}}} &
		{F1-score, AUROC} & {\textbf{Yes}} & {ED2 Accuracy} \\
		& {Fraction of 3 types} & {No} & {Algorithm-specific}  \\
		& {Rule width/\# of rules} & {\textbf{Yes}} & {ED2 Conciseness} \\ \hline
		{{SHAP~\cite{LundbergL17}}} & 
		{Visually on MNIST} & {No} & {Visual inspection}  \\ \hline
		{{Tree Reg.~\cite{WuHPZ2018}}} & 
		{AUROC} & {\textbf{Yes}} & {ED2 Accuracy}  \\ \hline
	\end{tabular}
	\caption{\small Metrics used in other ED works and applicability of the \system\ metrics}
	\label{tab:survey_measurement}
\end{table}
\endgroup

\section{Details on Pipeline Design}
\label{appendix:pipeline}

\minipa{1. Data Partitioning.}
This initial phase takes as input the 93 raw traces, as described in \S \ref{sec:anomalies}. It first performs simple data cleaning, e.g., replacing missing data with a default value. It then performs data selection and partitioning of the 93 traces according to the selected learning setting among the ones described in \S \ref{sec:benchmark}:
(i) In the {\em 1-App} learning settings (LS1 and LS3), we consider each Spark application, $i$, separately. The data selection step collects all the traces relating to $i$. The data partitioning step then takes all undisturbed traces of $i$ as training data, $D_{\text{train}}$, and all disturbed traces of $i$ as test data, $D_{\text{test}}$. These two datasets will be used to run the full pipeline, learning the model for $i$ and conducting the final evaluation. We repeat this process for each Spark application. 
(ii) In the {\em N-App} learning settings (LS2 and LS4), the data selection step collects all 93 traces. The data partitioning step then takes all  undisturbed traces as training data, $D_{\text{train}}$, and all disturbed traces as test data, $D_{\text{test}}$.
(iii) The above implementations correspond to the {\em Few-Examples} learning settings (LS3 and LS4). If the pipeline is configured to run under the  {\em Many-Examples} settings (LS2 and LS4), we further augment $D_{\text{train}}$, for {\em 1-App} and {\em N-App} learning, respectively, with an earlier segment of each relevant test trace (mostly normal data).

\minipa{7. ED Evaluation.} 
After processing each test trace, we obtain a set of anomalies with their corresponding explanations. We then collect the explanations from all the test traces to run the final ED evaluation.
To compute our metrics, we implemented for each ED method:
(i) The extraction function $G_\mathbf{A} (F_{t,w})$, used to extract the feature set appearing in a given explanation, required for conciseness and consistency. We support extraction functions for common forms of explanations, including logical formulas, linear models, and decisions trees. 
(ii) The subsampling procedure, applied to each anomaly $X_{t,w}$  when computing its ED1 stability and accuracy. If an ED method can explain an anomaly of arbitrary duration $w$, our sampling procedure performs a random split so that 80\% of the anomaly is used to construct its explanation (and its remaining 20\% is used along with neighboring normal data to compute its ED1 accuracy). For ED methods that can only explain anomalies of fixed size $s$, e.g., LIME~\cite{Ribeiro0G16}, we create samples of size $s$ such that their union entirely covers $X_{t,w}$.
The implementation of concordance and ED2 accuracy directly follows their definitions. 

\section{Details of Our Experimental Study}
\label{appendix:experiments}

All of our experiments were performed in a cluster of 20 compute nodes, each with 2 Intel{\scriptsize{\textsuperscript{\textregistered}}} Xeon{\scriptsize{\textsuperscript{\textregistered}}} Gold 6130 16-core processors, 768GB of memory, and 64TB disk. 

\subsection{Customized Feature Set}
\label{appendix:custom-set}

Our custom feature set was produced via the manual selection of relevant features resulting from domain knowledge and various visualizations. This feature set characterizes different aspects of a running application, with 9 features relating to the Spark application driver, 6 to the Spark executors, and 4 to the OS metrics. We give the complete list of features below: 

\begin{itemize}
	\item \textbf{\underline{Driver features}}
	\begin{itemize}
	\item Processing delay, scheduling delay and total delay of the last completed batch.
	\item First-order difference ($f_t := f_{t+1} - f_t$, simply called \emph{difference} in the following) in number of completed batches, received records and processed records since the start of the application.
	\item Difference in number of records in the last received batch. 
	\item Difference in total memory used.
	\item Difference in heap memory used by the JVM.
	\end{itemize}
	\item \textbf{\underline{Executor features}} -- All averaged across active Spark executors (before differencing).
	\begin{itemize}
	\item Difference in number of HDFS writing operations.
	\item Difference in CPU time and runtime.
	\item Difference in number of records read and written during shuffling operations. 
	\item Difference in heap memory used by the JVM.
	\end{itemize}
	\item \textbf{\underline{OS features}} -- All for each of the four nodes of the mini-cluster the applications were run on.
	\begin{itemize}
	\item Difference in global CPU percentage idle time.
	\end{itemize}
\end{itemize}
	
In the following, we list the feature names with their indices, as used in the explanations shown in Figure~\ref{fig:ed_example_all}. 

\begin{enumerate}[label={\arabic*}.]
	\item[0.] driver{\_}Streaming{\_}lastCompletedBatch{\_}processingDelay{\_}value
	\item driver{\_}Streaming{\_}lastCompletedBatch{\_}schedulingDelay{\_}value
	\item driver{\_}Streaming{\_}lastCompletedBatch{\_}totalDelay{\_}value
	\item 1{\_}diff{\_}driver{\_}Streaming{\_}totalCompletedBatches{\_}value
	\item 1{\_}diff{\_}driver{\_}Streaming{\_}totalProcessedRecords{\_}value
	\item 1{\_}diff{\_}driver{\_}Streaming{\_}totalReceivedRecords{\_}value
	\item 1{\_}diff{\_}driver{\_}Streaming{\_}lastReceivedBatch{\_}records{\_}value
	\item 1{\_}diff{\_}driver{\_}BlockManager{\_}memory{\_}memUsed{\_}MB{\_}value
	\item 1{\_}diff{\_}driver{\_}jvm{\_}heap{\_}used{\_}value
	\item 1{\_}diff{\_}node5{\_}CPU{\_}ALL{\_}Idle\%
	\item 1{\_}diff{\_}node6{\_}CPU{\_}ALL{\_}Idle\%
	\item 1{\_}diff{\_}node7{\_}CPU{\_}ALL{\_}Idle\%
	\item 1{\_}diff{\_}node8{\_}CPU{\_}ALL{\_}Idle\%
	\item 1{\_}diff{\_}avg{\_}executor{\_}filesystem{\_}hdfs{\_}write{\_}ops{\_}value
	\item 1{\_}diff{\_}avg{\_}executor{\_}cpuTime{\_}count
	\item 1{\_}diff{\_}avg{\_}executor{\_}runTime{\_}count
	\item 1{\_}diff{\_}avg{\_}executor{\_}shuffleRecordsRead{\_}count
	\item 1{\_}diff{\_}avg{\_}executor{\_}shuffleRecordsWritten{\_}count
	\item 1{\_}diff{\_}avg{\_}jvm{\_}heap{\_}used{\_}value
\end{enumerate}

\subsection{Details of AD Methods}
\label{appendix:ad-methods}

We next describe in more detail the methods we used for the normality modeling, outlier score derivation and threshold selection steps of AD modeling when conducting our experiments.
 
\textbf{LSTM}. Long Short-Term Memory (LSTM) \cite{DHochreiterS97} is a deep neural network structure designed for handling sequential data. In the context of AD, it is typically used as a forecasting-based method, where the outlier score of a given data point is derived from its forecasting error by the model. Like in a recent work~\cite{BontempsCML16}, our experiments used LSTM for AD by setting the outlier score of a test record as its relative forecasting error. Unlike in~\cite{BontempsCML16}, the produced outlier scores were however not further averaged but kept as is.

\textbf{AE}. An Autoencoder (AE) \cite{HS} is a deep neural network structure designed to map its input data to a corresponding latent representation in lower dimensional space (or \emph{coding}), by learning to accurately reconstruct it. In the context of AD, autoencoders are typically used as reconstruction-based methods, where the outlier score of a given data sample is derived from its reconstruction error by the model. In our experiments, we used a dense autoencoder architecture for AD by setting the outlier score of a test window as its reconstruction mean squared error (MSE), and the outlier score of a test record as the average score of its enclosing windows. 

\textbf{BiGAN}. Generative Adversarial Networks (GAN)~\cite{GoodfellowPMXWOCB14} learn to generate samples with similar characteristics as their training data, by training a generator network to map random noise in latent space to samples that appear realistic to a discriminator network. Bidirectional Generative Adversarial Networks (BiGAN)~\cite{DonahueKD17} are used to automatically learn the inverse mapping of the generator, through the additional training of an encoder network, that can be later used in pair with the generator to reconstruct some (real) input samples. As such, BiGAN can also be used in the context of AD as a reconstruction-based method, with an additional flexibility in deriving the outlier score of a test window, by leveraging the architectures of all three networks~\cite{abs180206222}. In our pipeline, we experimented with network architectures allowing the usage of LSTM networks to better handle time sequences, like done in~\cite{abs180904758}. The outlier score of a test window was defined as the average of its MSE by the (encoder, generator) pair and its feature loss by the discriminator, as defined in \cite{abs180206222}. Like for AE, the outlier score of a test record was defined as the average score of its enclosing windows.

\textbf{Threshold Selection}. Throughout our experiments, threshold selection was performed for an AD method based on the distribution of the outlier scores it assigned to the samples of $D_{\text{train}}^2$. More precisely, we experimented with threshold definitions of the form:

\[
\texttt{threshold} = S_1 + c \cdot S_2
\]

Where $S_1$ and $S_2$ are two descriptive statistics of the $D_{\text{train}}^2$ sample outlier scores, and $c$ is a constant we refer to as the \emph{thresholding factor}. We also allowed this computation to be performed twice, removing for the second iteration the outlier scores that were above the threshold found for the first iteration, in order to drop any obvious outliers that could prevent us from finding a suitable threshold.

For every AD evaluation in our experiments, the following threshold selection methods were evaluated for $c \in \left\{1.5, 2, 2.5, 3\right\}$: 

\begin{itemize}
    \item \textbf{STD} : $S_1 = \bar m$, $S_2 = s$, \emph{i.e.} sample mean and standard deviation.
    \item \textbf{MAD} : $S_1 = \text{median}, S_2 = \text{MAD}$, where $\text{MAD} = 1.4826 \cdot \text{median}(\lvert X - \text{median}(X) \rvert)$.
    \item \textbf{IQR} : $S_1 = Q_3$, $S_2 = Q_3 - Q_1 = \text{IQR}$.
\end{itemize}

This led to a total of 24 thresholding parameter combinations, corresponding to 24 final detection performances for each evaluated AD method, for which we reported the median values in Table~\ref{tab:custom-ls4-ad-detection}.

\subsection{Details of ED Methods}
\label{appendix:ed-methods}

We also integrated three recent ED methods into our pipeline: EXstream~\cite{ZDM2017} and MacroBase~\cite{macrobase:sigmod17} from the database community for outlier explanation in data streams, and LIME~\cite{Ribeiro0G16}, an influential ED method from the ML community. 

{\bf EXstream}~\cite{ZDM2017} takes as input an anomaly dataset and a normal, reference, dataset. It first calculates the entropy-based single-feature reward, which indicates the ability of a feature to separate normal from anomalous data, and then uses several heuristics to find a small set of high-reward features to build the explanation. One of its original steps was false positive filtering, pruning the features only incidentally correlated with abnormality. Since this step required generating extra labeled data based on user annotations, unavailable in our problem setting, we did not include it in our implementation.

{\bf MacroBase}~\cite{macrobase:sigmod17} considers the same input format as EXstream. It first calculates the risk ratio for each feature, and then for each combination of features based on a frequent itemset mining procedure. It also employs several optimization techniques to improve efficiency. %
Since it was originally designed for categorical features, we added the extra step of assigning categorical values to our numerical features via equal width binning in our implementation.

{\bf LIME}~\cite{Ribeiro0G16} was originally designed to explain a given prediction of a classifier, but also provides an interface for regression models and temporal inputs. Internally, LIME first leverages Lasso~\cite{efron_2004} to identify $k$ important features, and then uses a customized loss function to learn a new linear model approximating the original model's output locally. In our implementation, we used LIME's RecurrentTabularExplainer interface, and set $k=5$. 

\begin{figure*}[!b]
	\centering
	\begin{tabular}{lc}
		\hspace{-0.2in}
		\subfigure[\small{LSTM record-wise outlier scores}]
		{\label{fig:lm4-lstm-app1-cpu}\includegraphics[width=0.49\textwidth,height=2.0cm]{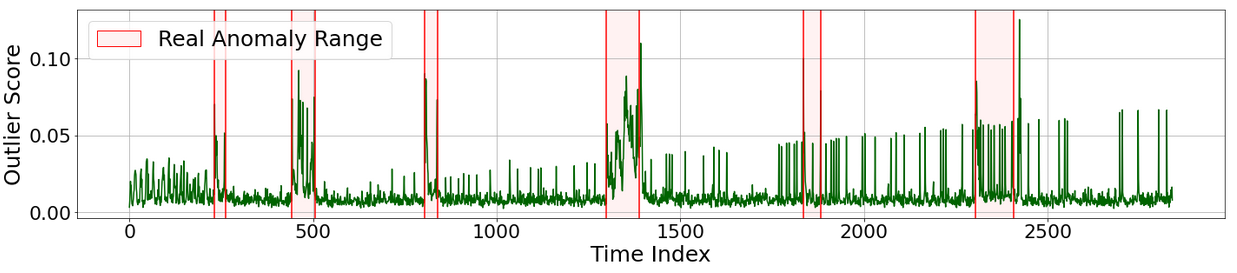}}
		&
		\subfigure[\small{AE record-wise outlier scores}]
		{\label{fig:lm4-ae-app1-cpu}\includegraphics[width=0.49\textwidth,height=2.0cm]{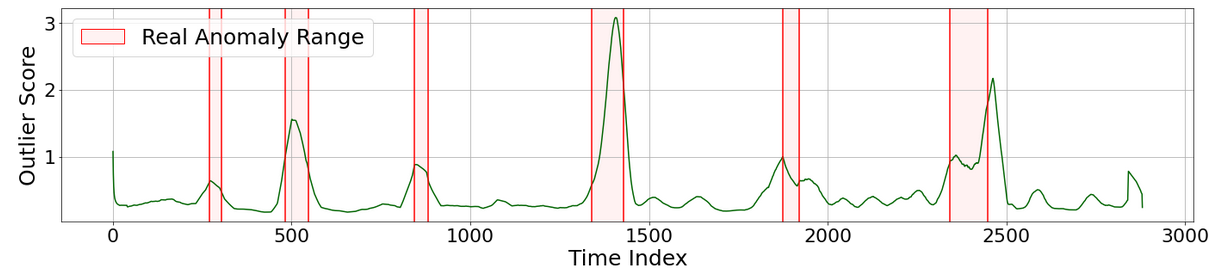}} \\ 
	\end{tabular}
\vspace{-0.2in}
\caption{\small Record-wise outlier scores on a T4 trace of Application 1 (FS$_\text{custom}$)}
\label{fig:lm4-app1-scores}
\end{figure*}

\subsection{Additional AD Evaluation Results}
\label{appendix:ad-results}

As shown in Figure~\ref{fig:lm4-app1-scores} for a T4 trace of Application 1, the AE method typically produces {\em smooth} record-wise outlier scores, by taking averages over overlapping windows, while the ones of LSTM often exhibit discontinuous {\em spikes}.
The outlier scores of the last two records were removed for LSTM, as their higher values visually packed down the others.

In the rest of this section, we describe the two additional AD experiments we conducted, respectively relating to the impact of feature extraction and learning setting on the AD performance. 

\minipa{Experiment 5 (AD2).}
In this experiment, we study the effect of feature extraction on the performance of the three methods, by comparing our custom feature selection (FS$_{\text{custom}}$) to applying PCA to the raw input features directly (FS$_{\text{pca}}$). From Table~\ref{tab:pca-ls4-ad2-global-separation}, we can see that using FS$_{\text{pca}}$ caused the global separation abilities of all methods to drop significantly (especially for LSTM), largely due to the fact that PCA was not able to preserve the right signals for detecting a major portion of our anomalies. 

When using FS$_{\text{pca}}$, we are not averaging across active executor features anymore, which likely plays a part in the fact that separation abilities for T6 were either maintained or increased for AE and BiGAN. Another aspect of FS$_{\text{pca}}$ is that it selects features based on their variance, which means that features tending not to vary much in normal scenarios, like scheduling delay or input rate, will be far less represented in the output space than features having naturally more activity, like CPU- or memory-related metrics. This likely explains why T4 separation ability was also maintained or increased for AE and BiGAN. Like previously hinted by remark R3, we can see this benchmark can be used to analyze the effect of constructing different feature spaces on the overall AD down to the detection of specific anomaly types. 

\begin{table}[t]
    \centering
    \small
    \begin{tabular}{|c|c|cccccc|}
    	\hline
    	 Method & Ave & \multicolumn{6}{c|}{AUPRC for Anomaly Types T1$\rightarrow$T6}\\ \hline \hline
   		\textbf{LSTM} & 0.14 & 0.15 & 0.14 & 0.09 & 0.14 & 0.21 & 0.12 \\
   		\textbf{AE} & 0.39 & 0.43 & 0.39 & 0.18 & 0.49 & 0.49 & 0.38 \\
   		\textbf{BiGAN} & 0.34 & 0.40 & 0.29 & 0.21 & 0.47 & 0.37 & 0.29 \\ \hline
    \end{tabular}
    \caption{\small Global separation results (FS$_{\text{pca}}$, AD2)}
    \label{tab:pca-ls4-ad2-global-separation}
\end{table}

\minipa{Experiment 6 (FS$_{\text{custom}}$, AD2).}
In this last experiment, we study the effect of using different learning settings for the three AD methods. Table~\ref{tab:custom-ls1:4-ad2-app-separation} reports their application-level separation abilities for each learning setting. From these, we see that AE and BiGAN followed the expected trend, with their average separation ability decreasing as we increase data variety and disable peeking at the beginning of test traces. On the contrary, LSTM seemed to perform better for LS2 and LS4 ({\em N-App} settings), which seems consistent with Experiment 3 showing that increasing the amount of training data for LSTM increases its performance, even when data variety is also increased. Although the changes in results appear as minor in these examples, we can see that training models with different learning settings can help draw some conclusions regarding their data needs and generalization ability ({\bf R8}).

\begin{table}[t]
    \centering
    \small
    \begin{tabular}{|c|c|c|cccccc|}
    	\hline
    	 LS & Method & Ave & \multicolumn{6}{c|}{AUPRC for Anomaly Types T1$\rightarrow$T6}\\ \hline \hline
   		\multirow{3}{*}{\textbf{LS1}} & 
   		\textbf{LSTM} & 0.43 & 0.55 & 0.35 & 0.46 & 0.33 & 0.58 & 0.32 \\
   		& \textbf{AE} & 0.58 & 0.72 & 0.41 & 0.67 & 0.54 & 0.71 & 0.42 \\
   		& \textbf{BiGAN} & 0.55 & 0.82 & 0.35 & 0.37 & 0.53 & 0.77 & 0.47 \\ \hline \hline
   		\multirow{3}{*}{\textbf{LS2}} & 
   		\textbf{LSTM} & 0.46 & 0.55 & 0.37 & 0.55 & 0.36 & 0.61 & 0.32 \\
   		& \textbf{AE} & 0.58 & 0.66 & 0.41 & 0.61 & 0.57 & 0.75 & 0.45 \\
   		& \textbf{BiGAN} & 0.54 & 0.72 & 0.33 & 0.65 & 0.56 & 0.68 & 0.31 \\ \hline \hline
   		\multirow{3}{*}{\textbf{LS3}} & 
   		\textbf{LSTM} & 0.43 & 0.58 & 0.34 & 0.44 & 0.33 & 0.58 & 0.30 \\
   		& \textbf{AE} & 0.57 & 0.72 & 0.40 & 0.64 & 0.53 & 0.69 & 0.44 \\
   		& \textbf{BiGAN} & 0.52 & 0.83 & 0.31 & 0.33 & 0.51 & 0.72 & 0.43 \\ \hline \hline
   		\multirow{3}{*}{\textbf{LS4}} & 
   		\textbf{LSTM} & 0.47 & 0.57 & 0.37 & 0.56 & 0.38 & 0.60 & 0.35 \\
   		& \textbf{AE} & 0.57 & 0.65 & 0.40 & 0.63 & 0.55 & 0.79 & 0.43 \\
   		& \textbf{BiGAN} & 0.52 & 0.81 & 0.36 & 0.25 & 0.54 & 0.69 & 0.48 \\ \hline
    \end{tabular}
    \caption{\small App-level separation results (LS1:4, FS$_{\text{custom}}$, AD2)}
    \label{tab:custom-ls1:4-ad2-app-separation}
\end{table}

\end{document}